\documentclass[10pt,journal,compsoc]{IEEEtran}
\usepackage[utf8]{inputenc}
\ifCLASSOPTIONcompsoc
  \usepackage[nocompress]{cite}
\else
  \usepackage{cite}
\fi

\usepackage{times}
\usepackage{epsfig}
\usepackage{graphicx}
\usepackage{amsmath}
\usepackage{bbold}
\usepackage{amssymb}
\usepackage{pifont}
\usepackage{bbm}
\usepackage{caption}
\usepackage{subcaption}
\usepackage[pagebackref=true,breaklinks=true,colorlinks,bookmarks=false]{hyperref}

\usepackage{review}

\usepackage{xspace}

\makeatletter
\DeclareRobustCommand\onedot{\futurelet\@let@token\@onedot}
\def\@onedot{\ifx\@let@token.\else.\null\fi\xspace}

\def\eg{\emph{e.g}\onedot} 
\def\ie{\emph{i.e}\onedot}

\def\wrt{w.r.t\onedot} 
\def\etal{\emph{et al}\onedot}
\makeatother

\newcommand{\sota}{state of the art }
\newcommand{\icl}{ICL}
\newcommand{\ours}{MiB }
\newcommand{\expandednick}{\textbf{M}odel\textbf{i}ng the \textbf{B}ackground for incremental learning in semantic segmentation}

\newcommand{\hypers}{hyper-parameters }
\newcommand{\hyper}{hyper-parameter }
\newcommand{\set}{\mathcal}
\newcommand{\con}{\mathtt}
\DeclareMathOperator*{\argmax}{arg\,max} %
\newcommand{\real}{{\rm I\!R}}

\setrevision{0}
\newcommand{\comRev}[1]{{#1}}

\newcommand{\myparagraph}[1]{\vspace{4pt}\noindent\textbf{#1}}

\begin{document}

\begin{titlepage}
\null
\vfill
\renewcommand{\fboxsep}{10pt}
\centering
\fbox{\Large\begin{minipage}{\columnwidth}
\textbf{Disclaimer:}

This work has been accepted for publication in the IEEE Transactions on Pattern Analysis and Machine Intelligence:\vspace{4pt}
\newline
doi:     10.1109/TPAMI.2021.3133954
\newline
link:    https://ieeexplore.ieee.org/document/9645239/
\newline
\newline
\textbf{Copyright:} 
\newline
\copyright~2021 IEEE. Personal use of this material is permitted. Permission from IEEE must be obtained for all other uses,  in  any  current  or  future  media,  including  reprinting/  republishing  this  material  for  advertising  or promotional purposes, creating new collective works, for resale or redistribution to servers or lists, or reuse of any copyrighted component of this work in other works.
\newline
\end{minipage}}
\vfill
\clearpage
\end{titlepage}

\title{Modeling the Background for Incremental and Weakly-Supervised Semantic Segmentation}

\author{Fabio~Cermelli,~\IEEEmembership{}
        Massimiliano~Mancini,~\IEEEmembership{}
        Samuel~Rota~Bul\'o,~\IEEEmembership{}
        Elisa~Ricci~\IEEEmembership{}
        and Barbara~Caputo~\IEEEmembership{}%
\IEEEcompsocitemizethanks{\IEEEcompsocthanksitem F. Cermelli and B.Caputo are with DAUIN Department of Control and Computer  Engineering  of  Politecnico  di  Torino, Turin, Italy. (Email: fabio.cermelli@polito.it, barbara.caputo@polito.it)
\IEEEcompsocthanksitem M. Mancini is with Cluster of Excellence "Machine Learning", University of Tübingen, Germany. (Email: massimiliano.mancini@uni-tuebingen.de)
\IEEEcompsocthanksitem S. Rota Bul\'o is with Mapillary Research, Graz, Austria.
\IEEEcompsocthanksitem F.Cermelli and B. Caputo are with Italian Institute of Technology, Turin, Italy.
\IEEEcompsocthanksitem E. Ricci is with Fondazione Bruno Kessler, Trento, Italy. (Email: eliricci@fbk.eu)
\IEEEcompsocthanksitem E. Ricci is with Department of Information Engineering and Computer Science, University of Trento, Trento, Italy}%
}

\IEEEtitleabstractindextext{%
\begin{abstract}
Deep neural networks have enabled major progresses in semantic segmentation. However, even the most advanced neural architectures suffer from important limitations. First, they are vulnerable to catastrophic forgetting, i.e. they perform poorly when they are required to incrementally update their model as new classes are available. Second, they rely on large amount of pixel-level annotations to produce accurate segmentation maps. To tackle these issues, we introduce a novel incremental class learning approach for semantic segmentation taking into account a peculiar aspect of this task: since each training step provides annotation only for a subset of all possible classes, pixels of the background class exhibit a semantic shift. Therefore, we revisit the traditional distillation paradigm by designing novel loss terms which explicitly account for the background shift. Additionally, we introduce a novel strategy to initialize classifier’s parameters at each step in order to prevent biased predictions toward the background class. Finally, we demonstrate that our approach can be extended to point- and scribble-based weakly supervised segmentation, modeling the partial annotations to create priors for unlabeled pixels.
We demonstrate the effectiveness of our approach with an extensive evaluation on the Pascal-VOC, ADE20K, and Cityscapes datasets, significantly outperforming state-of-the-art methods. 
\end{abstract}

}

\maketitle

\IEEEdisplaynontitleabstractindextext
\IEEEpeerreviewmaketitle
\IEEEraisesectionheading{\section{Introduction}}

 \IEEEPARstart{T}{he} goal of semantic segmentation \cite{long2015fully} is to correctly predict the semantic label associated to each pixel in an image. %
In the last years, thanks to the emergence of deep neural networks and to the availability of large-scale human-annotated datasets \cite{pascal-voc-2012,zhou2017scene}, the state of the art in this task 
has improved significantly \cite{long2015fully, chen2018encoder, zhao2017pyramid, lin2017refinenet, zhang2018exfuse}.
Current approaches   
are based %
on Fully Convolutional Networks (FCNs) \cite{long2015fully} and mostly differ from the strategies used to combine multiscale representations \cite{lin2017refinenet,zhang2018exfuse}, to model spatial dependencies and contextual cues \cite{chen2017rethinking, chen2017deeplab, chen2018encoder} or to integrate attention models \cite{chen2016attention}.

\begin{figure}[t]
    \centering
    \includegraphics[width=\linewidth,trim=6cm 0 0 0, clip]{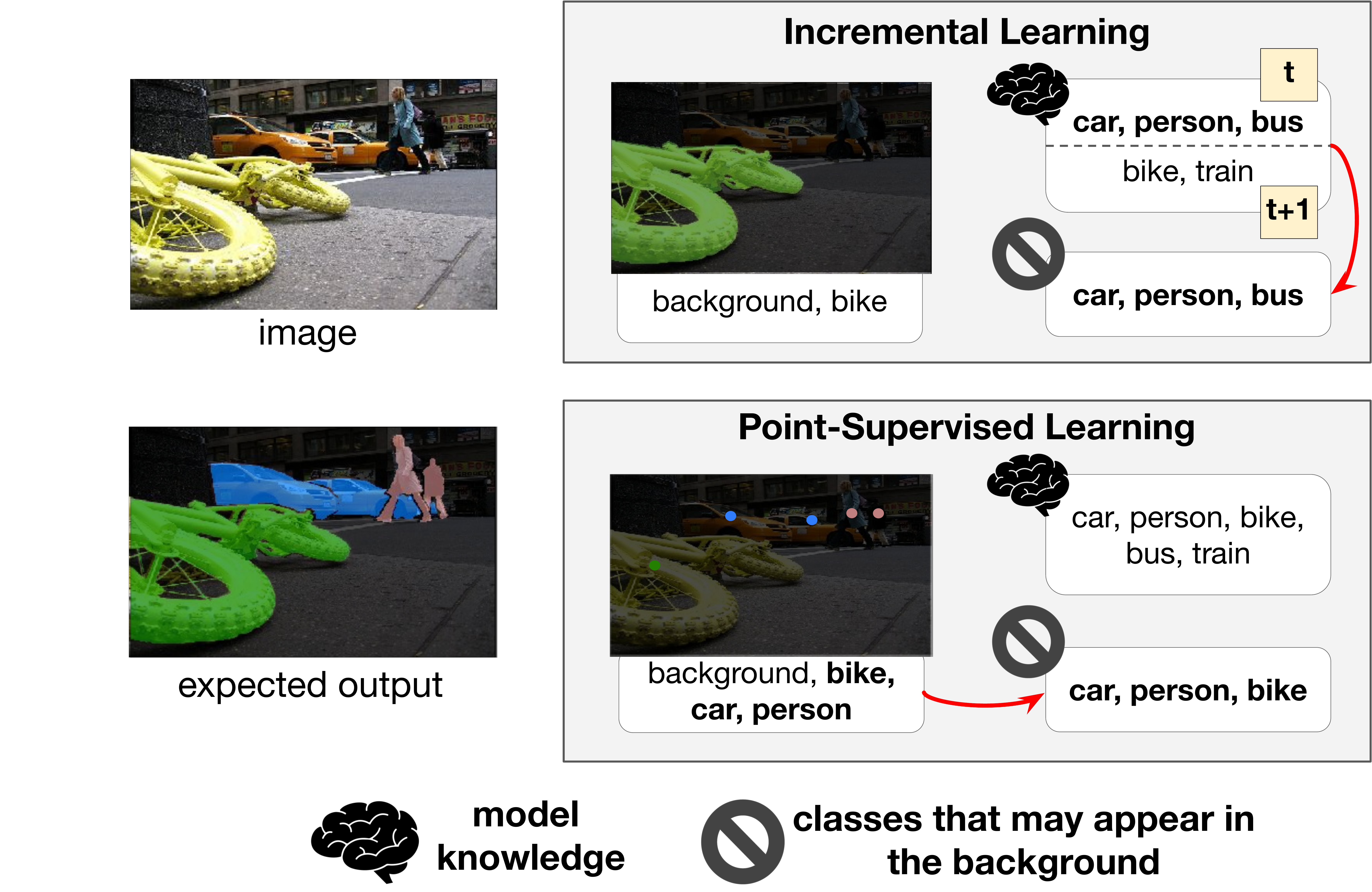}
    \caption{The figure %
    depicts the content of the background pixels in case of partial labels. In incremental learning (top), since we have labels only for pixels of novel classes in the current training step, the background may contain pixels of the old ones. %
    In point-supervised learning, every class with at least one annotated point in the image is also present in the background. Image taken from the Pascal-VOC dataset \cite{pascal-voc-2012}.} %
    \label{fig:teaser}
    \vspace{-10pt}
\end{figure}

Despite their effectiveness, semantic segmentation models need a large amount of images with paired pixel-level annotations during training, which are extremely costly to collect. This can be overcome 
by training semantic segmentation models %
with weaker forms of supervisions, such as image-level labels \cite{kolesnikov2016seed} and points \cite{bearman2016s}. Still, both fully-supervised and weakly-supervised learning (WSL) algorithms assume that the annotated data for all the semantic categories the model will be asked to recognize should be available beforehand. %
This assumption rarely holds in many practical applications; 
it would be desirable to dispose of semantic segmentation models able to continuously incorporate information about novel categories, while being able to retain knowledge about the previous classes. %
In this paper we study the problem of semantic segmentation in an incremental class learning (\icl) scenario \cite{rebuffi2017icarl}, \ie we aim to build a deep model %
able to incrementally learn new categories whilst preserving good performance on the old ones avoiding {catastrophic forgetting} \cite{mccloskey1989catastrophic}.

Our approach is inspired by previous \icl\ methods on image classification \cite{li2017learning,rebuffi2017icarl,castro2018end}, which address catastrophic forgetting through knowledge distillation \cite{hinton2015distilling}. However, 
here we show that a naive application of previous knowledge distillation strategies would not suffice in our setting. The reason is that none of these approaches take explicitly into account the evolving semantics of the background class among different training steps, a problem that we called \textit{background shift}. Indeed, since we have only partial annotations in each training step, unlabeled/background pixels might belong to some of the classes we have previously learned and even to classes we will learn in the future. For instance (Fig.~\ref{fig:teaser}, top), we might have learned the classes \textit{car}, \textit{person} and \textit{bus} at a previous learning step $t$, and at step $t+1$ we learn new classes (e.g., \textit{bike} and \textit{train}). Since in the current training step we have annotations only for new classes, the background might contain pixels of old classes as well. Note that this problem is peculiar to semantic segmentation. To overcome this issue, we revisit the classical distillation-based framework in \cite{li2017learning} by introducing two novel loss terms to properly account for the semantic distribution shift within the background class. 
Our approach is based on a simple principle: each background pixel in the ground-truth might contain either the background or one of the categories whose annotation is missing for the current image.
This means that, in the incremental learning setting, we consider pixels labeled as background at a given learning step to contain either the background or any of the previously seen classes. A similar reasoning can be applied for the distillation loss, in a symmetric manner. 
We extensively evaluate our method on three datasets, Pascal-VOC \cite{pascal-voc-2012}, ADE20K \cite{zhou2017scene} and Cityscapes \cite{cordts2016cityscapes} showing that our approach, coupled with a novel classifier initialization strategy, largely outperform traditional \icl\ methods. 

Finally, we show how our \icl\ approach easily extends to other \comRev{partially-annotated} scenarios, such as weakly supervised semantic segmentation with point \comRev{or scribble} supervision. In this setting, we consider %
\comRev{non-annotated pixels} containing either the actual background or any of the weakly annotated classes in the current image. {As an example, in Fig.\ref{fig:teaser} (bottom), we have point-level annotations for \textit{bike}, \textit{car} and \textit{person}. The definition of the setting entails that \comRev{non-annotated} pixels might contain either the background or one of the three classes above, but it does not contain other classes (e.g., \textit{train} and \textit{bus}) without any annotation in the current image.} We encode this prior in the standard cross-entropy loss and we benchmark this approach in semantic segmentation in Pascal-VOC \comRev{using point \cite{bearman2016s} and scribble \cite{lin2016scribblesup} supervision, showing performance superior or comparable} to the state of the art. We also evaluate our method on another scenario, scene parsing \comRev{with point supervision}, where the background is not present but unlabeled pixels might still contain any of the classes with at least one point in the current image. Experiments on ADE20k \cite{qian2019weakly} demonstrate the effectiveness of our approach in this scenario.

\noindent To summarize, the main contributions of this paper are as follows:
\begin{itemize}
     \item We identify the problem of semantic shift of the background class arising in incremental class learning for semantic segmentation.
    \item We revisit standard \icl\ approaches with a novel objective function that is applied both to a cross-entropy and a standard distillation loss. Coupled with a specific classifier initialization strategy, our approach greatly alleviates the catastrophic forgetting and the semantic shift of the background class, leading to the state of the art.%
    \item We benchmark our approach over several previous \icl\ methods on three popular semantic segmentation datasets, considering different experimental settings. We hope that our results will serve as a reference for future works in incremental learning in semantic segmentation. %
    \item We show how the same approach can be applied to the task of WSL using point \comRev{or scribble} supervision, achieving state-of-the-art results in \comRev{three} different experimental settings.
\end{itemize}

This paper extends our earlier work \cite{cermelli2020modeling} in many aspects. 
\addnote[intro:novelty]{1}{
In particular, we demonstrate that the key idea behind modeling the background, \ie considering unlabeled/background pixels as belonging to any class in a specific set built through known priors (\ie old classes), can be extended to any partially-annotated scenario, such as semantic segmentation with point or scribble supervision.
Experiments demonstrate that our approach is very effective even in \comRev{these new tasks}, confirming that its underlying idea of modeling the semantic of the background class is general and it is applicable on multiple tasks consisting of noisy or partial annotations.} 
Finally, we expanded our experimental evaluation on \icl\ considering other challenging scenarios in the Cityscapes dataset \cite{cordts2016cityscapes}. We also provide a more comprehensive review of related works, including the weakly-supervised learning literature.

The rest of this paper is organized as follows. We first introduce related work in Section \ref{sec:related} and then describe our \icl\ method and its extension to tackle WSL with \comRev{weak} supervision in Section \ref{sec:method}. The results of our approach on \icl\ and WSL are provided 
in Section \ref{sec:experiments}. We conclude the paper in Section \ref{sec:conclusions}.

The code is available at \url{https://github.com/fcdl94/MiB}.

\section{Related Works}
\label{sec:related}
\myparagraph{Semantic Segmentation.}
Deep learning has enabled great advancements in semantic segmentation \cite{long2015fully, chen2018encoder, zhao2017pyramid, lin2017refinenet, zhang2018exfuse}. State-of-the-art methods are based on Fully Convolutional Neural Networks \cite{long2015fully, badrinarayanan2017segnet} and use different %
strategies to condition pixel-level annotations on their global context, \eg using multiple scales \cite{zhao2017pyramid,lin2017refinenet,chen2017rethinking, chen2017deeplab, zhang2018exfuse, chen2018encoder} and/or modeling spatial dependencies \cite{chen2017rethinking,ghiasi2016laplacian}.  %
The vast majority of semantic segmentation methods considers an offline setting, \ie they assume that training data for all classes is available beforehand. 
To our knowledge, the problem of \icl\ in semantic segmentation has been addressed only in
\cite{ozdemir2018learn,ozdemir2019extending,tasar2019incremental,michieli2019incremental}. Ozdemir \etal \cite{ozdemir2018learn,ozdemir2019extending} describe an \icl\ approach for medical imaging, extending a standard image-level classification method \cite{li2017learning} to segmentation and devising a strategy to select relevant samples of old datasets for rehearsal. Tasar 
\etal \cite{tasar2019incremental} proposed a similar approach for segmenting remote sensing data. %
Differently, Michieli \etal
\cite{michieli2019incremental} consider \icl\ for semantic segmentation 
in a particular setting where labels are provided for old classes while learning new ones. Moreover, they assume the novel classes to be never present as background pixels in previous learning steps. These assumptions strongly limit the applicability of their method. %

Here we propose a more principled formulation of the \icl\ problem in semantic segmentation. %
In contrast with previous works, we do not limit our analysis to medical \cite{ozdemir2018learn} or remote sensing data \cite{tasar2019incremental} and we do not impose any restrictions on how the label space should change across different learning steps \cite{michieli2019incremental}.  Moreover, 
we are the first to provide a comprehensive  experimental evaluation of \sota \icl\ methods on commonly used semantic segmentation benchmarks %
and to explicitly introduce and tackle the semantic shift of the background class, %
a problem recognized but largely overseen by previous works \cite{michieli2019incremental}. Our strategy 
can be applied in different scenarios with partial annotations, such as weakly supervised learning.

\myparagraph{Incremental Learning.}
The problem of catastrophic forgetting \cite{mccloskey1989catastrophic} has been extensively studied for image classification tasks \cite{de2019continual}. 
{Previous works can be grouped in three categories \cite{de2019continual}: replay-based \cite{rebuffi2017icarl, castro2018end, shin2017continual, hou2019learning, wu2018memory, ostapenko2019learning}, regularization-based \cite{kirkpatrick2017overcoming,chaudhry2018riemannian,zenke2017continual,li2017learning, dhar2019learning}, and parameter isolation-based \cite{mallya2018packnet, mallya2018piggyback, rusu2016progressive}.
In replay-based methods, examples of previous tasks are either stored \cite{rebuffi2017icarl, castro2018end, hou2019learning, wu2019large} or generated \cite{shin2017continual, wu2018memory, ostapenko2019learning} and then replayed while learning the new task. %
Parameter isolation-based methods \cite{mallya2018packnet, mallya2018piggyback, rusu2016progressive} assign a subset of the parameters to each task to prevent forgetting.} %
Regularization-based methods can be divided in prior-focused and data-focused. 
The former \cite{zenke2017continual, chaudhry2018riemannian, kirkpatrick2017overcoming, aljundi2018memory} define knowledge as the parameters value, constraining the learning of new tasks by penalizing changes of important parameters for old ones. 
The latter \cite{li2017learning, dhar2019learning,fini2020online} exploits distillation \cite{hinton2015distilling} and use the distance between the activations produced by the old network and the new one as a regularization term to prevent catastrophic forgetting. %

Despite these progresses, very few works have gone beyond image-level classification. A first work in this direction is \cite{shmelkov2017incremental} which considers \icl\ in object detection %
proposing a distillation-based method adapted from \cite{li2017learning} for tackling novel class recognition and bounding box proposals generation. 
In this work we also take a similar approach to \cite{shmelkov2017incremental} and we resort on distillation. However, here we specifically propose to address the problem of modeling the background shift which is peculiar of the semantic segmentation setting. %

\myparagraph{Weakly Supervised Learning.}
The significant burden of requiring annotations for each pixel of an image has lead to several research efforts toward building semantic segmentation models using cheaper (but weaker) annotations. Under this perspective, different types of annotation has been explored, such as image-level labels \cite{lee2019ficklenet, huang2018weakly, kolesnikov2016seed,sun2020mining}, bounding boxes \cite{dai2015boxsup, papandreou2015weakly,khoreva2017simple}, scribbles \cite{lin2016scribblesup,tang2018normalized} and points \cite{bearman2016s,qian2019weakly}.

\addnote[rel:weakly]{1}{Image-level labels only provide information about which classes are contained in the image, without any hint on their locations. Most approaches in this direction \cite{kolesnikov2016seed, oh2017exploiting, huang2018weakly, ahn2018learning, ahn2019weakly, lee2019ficklenet, sun2020mining, araslanov2020single, chang2020weakly} aim to generate pixel-wise pseudo-labels obtaining and refining an initial localization map, which is often a class activation maps (CAM) \cite{zhou2016cam, selvaraju2017gradcam} obtained from an image-level classifier. %
\cite{kolesnikov2016seed} introduced the idea to use CAMs as a seed for weak localization of the objects, expanding the object prediction based on the information provided by image-level labels and constraining the segmentation masks with CRF-based object boundaries. %
On the intuition that better localization cues may further improve performances, subsequent works proposed to refine the localization priors. In \cite{huang2018weakly} the seeded region growing algorithm \cite{adams1994seeded} is adapted to extend the prior, \cite{ahn2018learning} modeled the pixel similarity from the initial CAMs and employed random walk to propagate the class labels, \cite{lee2019ficklenet} extracted multiple class activation maps using different combinations of the image feature obtained with dropout \cite{srivastava2014dropout}, \cite{sun2020mining} exploited the cross-image semantic relation and \cite{chang2020weakly} adopted consistency regularization to improve the localization seed.
}

A stronger form of weak annotations are bounding boxes \cite{dai2015boxsup}, providing information about the classes in the image, their location and dimensions. In this scenario, various approaches explore the use of region proposal methods to refine the candidate object masks \cite{dai2015boxsup,khoreva2017simple}. To this extent, \cite{dai2015boxsup} uses multi-scale combinatorial grouping (MCG) \cite{arbelaez2014multiscale}, refining the masks in an iterative process involving the ground-truth bounding boxes and the network predictions. Similarly, in \cite{khoreva2017simple} the segmentation masks are refined using GrabCut \cite{rother2004grabcut}, MCG and the network predictions.

\addnote[rel:weakly-scribble]{1}{Finally, cheaper than bounding boxes are scribbles \cite{lin2016scribblesup} or points \cite{bearman2016s}. 
Scribble annotations provide a strong localization information and are very fast to collect, providing a class for each scribble. In this scenario, \cite{lin2016scribblesup} first proposed to expand the scribble supervision by dividing pixels into super-pixels and exploiting pixel-similarity as additional source of supervision. Differently, \cite{tang2018normalized, tang2018regularized} integrates graphical models (\eg, graph cut or dense CRFs) %
{into regularization losses  during training, forcing the model to produce consistent outputs on similar pixels. Recently, \cite{wang2019boundary} proposed to use two additional sub-networks to fully exploit scribble-annotation: one sub-network refines the model's output with an iterative up-sampling while the other performs boundary prediction to obtain more precise segmentation results.}}

\comRev{Point supervision is more challenging since it provides only one point for each instance in the image.} To solve this problem, in \cite{bearman2016s} %
the authors propose to use three main components: (i) an image-level prior to predict which objects are present in the image, (ii) a partial cross-entropy on the labeled points, and (iii) an objectness prior, extracted from a shallow model, which helps in differentiating background and foreground pixels.
In \cite{qian2019weakly} the authors propose a method using point supervision for the task of scene parsing, where a model is asked to segment both objects and stuffs. The authors propose to use the partial cross entropy coupled with a distance metric regularization, forcing pixels of the same classes to produce similar feature vectors. %

In this work we focus on \comRev{weak supervision with points and scribbles}, both in object segmentation and scene parsing settings. We show how our simple loss formulation considering the uncertainty on unlabeled pixels produces a boost on the performance of the standard partial cross-entropy adopted by multiple works, achieving state-of-the-art results in both scenarios.

\section{Modeling the Uncertainty in Semantic Segmentation}
\label{sec:method}
\subsection{Problem Definition}
The goal of semantic segmentation is to produce a model capable of assigning a class for each pixel of a given input image. 
Let us denote as $\mathcal{X}$ the input space (\ie the image space) and, without loss of generality, let us assume that each image $x\in\mathcal{X}$ is composed by a set of pixels $\set I$ with constant cardinality $|\set I|=N$. 
The output space is defined as $\mathcal{Y}^{N}$, with the latter denoting the product set of $N$-tuples with elements in a label space $\mathcal{Y}$. Given an image $x$ the goal of semantic segmentation is to assign each pixel $x_i$ of $x$ a label $y_i \in \mathcal{Y}$, representing its semantic class.
{
The mapping is realized by learning a model $f_{\theta}$ with parameters $\theta$ from the image space $\mathcal{X}$ to a pixel-wise class probability vector, \ie $f_{\theta} : \mathcal{X} \mapsto \mathcal{\real}^{N \times |\mathcal{Y}|}$.
To learn the mapping, it is provided a training set $\mathcal{T} \subset \mathcal{X} \times (\set Y \cup \con u)^N$, where $\con u$ indicates pixels that are not labeled, either because they contain objects which are not of interest or the labeling is partial. 
The output segmentation mask is obtained as $y^* = \{ \argmax_{c\in\mathcal{Y}} f_{\theta}(x)[i,c]\}_{i=1}^{N}$, where $f_{\theta}(x)[i,c]$ is the probability for class $c$ in pixel $i$. In the following, we will indicate the probability for class $c$ in pixel $i$  as $q_x(i,c) = f_{\theta}(x)[i,c]$.
}

{
\subsection{Learning from the Unknown}
\label{sec:method-unknown}
Commonly, in the training procedure of semantic segmentation, unlabeled pixels are discarded from the loss computation since it is believed that they do not bring information or it is not known what loss function should be minimized on them. However, we argue that, if it is possible to make assumptions on the classes they belong to, these pixels carry useful information that can be used in the training procedure. In particular, denoting with $\mathcal{U} \subseteq \mathcal{Y}$ the set of classes to whom unlabeled pixels might belong to, we can define a loss function on an image $x$, with label $y$ as:
\begin{equation} \label{eq:unknown}
 \ell(x, y) = - \sum_{i \in \set I} \log p_x(i,y_i) \,,
\end{equation}
where $y_i$ is the ground truth label associated to pixel $i$ and $p_x$ is computed as follow:
\begin{equation} \label{eq:cases-unknown}
    p_x(i,c) = \begin{cases}
      {q}_x(i,c)\;\;& \text{if}\ c\neq \con u\\
      \sum_{k\in \set U}{q}_x(i,k)\;\;& \text{if}\ c=\con u\,.
    \end{cases}
\end{equation}

The idea behind Eq.~\eqref{eq:unknown} and Eq.~\eqref{eq:cases-unknown} is that unlabeled pixels should provide a positive feedback for all the semantic classes they might contain. Being simple and general, this loss allows to exploit the information provided by both the labeled pixels (it degenerates to the standard cross entropy when there are no unlabeled pixels) and the unlabeled ones, through the prior that we have on their semantic content.

In the following, we first show this idea can be effectively used to address the background shift problem of \icl\ in semantic segmentation. \comRev{Next, we extend this idea to weakly-supervised learning, showing that modeling the unlabeled pixels is beneficial to improve the final performance.}
}

\section{Incremental Learning in Semantic Segmentation}
\begin{figure*}
    \centering
    \includegraphics[width=0.95\textwidth]{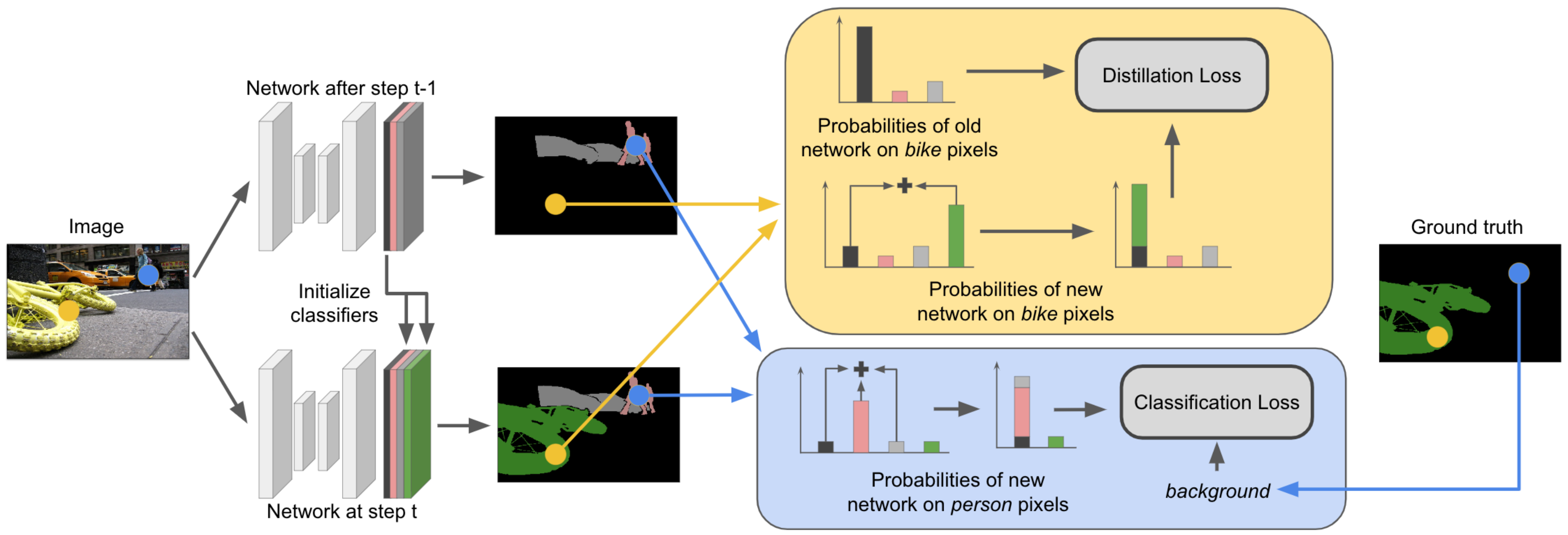}
    \caption{Overview of our method. At learning step $t$ an image is processed by the old (top) and current (bottom) models, mapping the image to their respective output spaces. As in standard \icl\ methods, we apply a cross-entropy loss to learn new classes (blue block) and a distillation loss to preserve old knowledge (yellow block). In this framework, we model the semantic changes of the background 
 across different learning steps by (i) initializing the new classifier using the weights of the old background one (left), (ii) comparing the pixel-level background ground truth in the cross-entropy with the probability of having either the background (black) or an old class (pink and grey bars) and (iii) relating the background probability given by the old model in the distillation loss with the probability of having either the background or a novel class (green bar). Image taken from the Pascal-VOC dataset \cite{pascal-voc-2012}. }
    \label{fig:method}
\end{figure*}

\label{sec:incremental}
In the \icl\ setting, training is realized over multiple phases, called \textit{learning steps}, and each step introduces novel categories to be learnt. In other terms, during the $t_{\text{th}}$ learning step, the previous label set $\set Y^{t-1}$ is expanded with a set of new {classes} $\set C^t$, yielding a new label set $\set Y^t=\set Y^{t-1}\cup\set C^{t}$.
{At learning step $t$ we are provided with a training set $\mathcal{T}$ which only contains labels for pixels of novel classes while all the other pixels of the image are unlabeled. However, ignoring these unlabeled pixels will prevent the model to learn the boundary of novel classes. For this reason, we decide to assign to the unlabeled pixels a special class, \ie the background class $\con b$. The background class is the only class which is shared by multiple learning steps and it is included in any label and class set, \ie $\con b \in \set C^t$ for any step $t$. 
The learning is then performed using the current training set $\mathcal{T}^t \subset \mathcal{X}\times (\set {C}^{t})^N$ in conjunction to the previous model $f_{\theta^{t-1}}:\set X\mapsto \real^{N\times |\set Y^{t-1}|}$ to obtain an updated model $f_{\theta^{t}}:\set X\mapsto \real^{N\times |\set Y^{t}|}$.
}
As in standard \icl, in this paper we assume the sets of labels $\mathcal{C}^t$ that we obtain at the different learning steps to be disjoint, except for the special background class $\con b$.

\subsection{Modeling the Background}
\label{sec:our-method}
A naive approach to address the \icl\ problem consists in retraining the model $f_{\theta^t}$ on each set $\mathcal{T}^t$ sequentially. When the predictor $f_{\theta^t}$ is realized through a deep architecture, this corresponds to fine-tuning the network parameters on the training set $\mathcal{T}^t$ initialized with the parameters $\theta^{t-1}$ from the previous stage. This approach is simple, but it leads to catastrophic forgetting. Indeed, when training using $\mathcal{T}^t$ no samples from the previously seen object classes %
are provided. This biases the new predictor $f_{\theta^t}$ %
towards the novel set of categories in $\mathcal{C}^t$ to the detriment of the classes from the previous sets. In the context of \icl\ for image-level classification, a standard way to address this issue is coupling the supervised loss on $\mathcal{T}^t$ with a regularization term, either taking into account the importance of each parameter for previous tasks \cite{kirkpatrick2017overcoming,shin2017continual}, or by distilling the knowledge using the predictions of the old model $f_{\theta^{t-1}}$ \cite{li2017learning,rebuffi2017icarl,castro2018end}. We take inspiration from the latter solution to initialize the overall objective function of our problem. In particular, we  minimize a loss function of the form:
\begin{equation}
   \label{eq:obj-general}
    \mathcal{L}(\theta^t)= \frac{1}{|\mathcal{T}^t| %
    }\sum_{(x,y)\in\mathcal{T}^t}
    \left(\ell^{\theta^t}_{ce}(x,y) + \lambda \ell^{\theta^t}_{kd}(x) \right)
\end{equation}
where $\ell_{ce}$ is a standard supervised loss (\eg cross-entropy loss), $\ell_{kd}$ is the distillation loss and $\lambda>0$ is a \hyper balancing the importance of the two terms. 

As stated at the beginning of the Sec. \ref{sec:incremental}, differently from standard \icl\ settings considered for image classification problems, in semantic segmentation we have that two different label sets $\mathcal{C}^s$ and $\mathcal{C}^u$ share the common background class $\mathtt{b}$. %
However, the distribution of the background class changes across different incremental steps. In fact, background annotations given in $\set T^t$ refer to classes not present in $\set C^t$, that might belong to the set of seen classes $\set Y^{t-1}$ and/or to still unseen classes \ie $\set C^{u}$ with $u>t$. 
{In the following, we show how to account for the semantic shift of the the background class by revisiting standard choices for the general objective defined in Eq.~\eqref{eq:obj-general} with our formulation in \ref{sec:method-unknown}. }

\myparagraph{Revisiting Cross-Entropy Loss. } In Eq.~\eqref{eq:obj-general}, a possible choice for $\ell_{ce}$ is the standard cross-entropy loss computed over all image pixels: 
\begin{equation}
   \label{eq:CE}
 \ell^{\theta^t}_{ce}(x,y) = - \frac{1}{|\set I|}\sum_{i \in\set I}\log q_x^t(i,y_i)\,,
\end{equation} 
where $q_x^t$ is the output of the model at the training step $t$, \ie $q_x^t(i,c) = f_{\theta_t}(x)[i,c]$

The problem with Eq.~\eqref{eq:CE} is that the training set $\mathcal{T}^t$ we use to update the model only contains information about novel classes in $\mathcal{C}^t$. However, the unlabeled pixels in $\mathcal{T}^t$, that are assigned to the background class, might include also pixels associated to the previously seen classes in $\set Y^{t-1}$. We argue that, without explicitly taking into account this aspect, the catastrophic forgetting problem would be even more severe. In fact, we would drive our model to predict the background label $\con b$ for pixels of old classes, further degrading the capability of the model to preserve semantic knowledge of past categories. {To avoid this issue, we propose to replace the cross-entropy loss in Eq.~\eqref{eq:CE} %
with the loss function in Eq.~\eqref{eq:unknown}, by considering $\set U = \set Y^{t-1}$}. Therefore, we define: 
\begin{equation}
  \label{eq:our-CE}
    \ell^{\theta^t}_{ce}(x,y) = -\frac{1}{|\set I|}\sum_{i \in\set I}\log p_x^t(i,y_i)\,,
 \end{equation}
 where:
\begin{equation}
    \label{eq:cases-ce}
     p_x^t(i,c) = \begin{cases}
      {q}_x^t(i,c)\;\;& \text{if}\ c\neq\mathtt{b}\\
      \sum_{k\in\mathcal{{Y}}^{t-1}}{q}_x^t(i,k)\;\;& \text{if}\ c=\mathtt{b}\,.
    \end{cases}
\end{equation}

Our intuition is that by using Eq.\eqref{eq:our-CE} we can update the model to predict the new classes and, at the same time, account for the uncertainty over the actual content of the background class. In fact, in Eq.~\eqref{eq:our-CE} the background class ground truth is not directly compared with its probabilities ${q}_x^t(i,\con b) $ obtained from the current model $f_{\theta^t}$, but with the probability of having \textit{either an old class or the background}.
{A schematic representation of this procedure is depicted in Fig.~\ref{fig:method} (blue block).}
 
It is worth noting that the alternative of ignoring the unlabeled pixels within the cross-entropy loss is a worse solution than considering them as background. In fact, this would not allow the model to correctly update its classifier to update its representation of the background and to learn the boundary of novel objects. Moreover, it would not allow to exploit the information that new images might contain about old classes. 

\myparagraph{Revisiting Distillation Loss.} In the context of incremental learning, distillation loss \cite{hinton2015distilling} is a common strategy to transfer knowledge from the old model $f_{\theta^{t-1}}$ into the new one, preventing catastrophic forgetting. 
Formally, a standard choice for the distillation loss $\ell_{kd}$ is:
\begin{equation}
    \label{eq:std-distill}
     \ell^{\theta^t}_{kd}(x,y) = \frac{1}{|\set I|}\sum_{i \in\set I}\sum_{c \in\set Y^{t-1}} q_x^{t-1}(i,c)\log \hat{p}_x^{t}(i,c)\,,
\end{equation}
where $\hat{p}_x^{t}(i,c)$ is defined as the probability of class $c$ for pixel $i$ given by $f_{\theta^t}$ but re-normalized across all the classes in $\set Y^{t-1}$ \ie:
\begin{equation}
    \label{eq:cases-kd}
    \hat{p}_x^{t}(i,c)= \begin{cases}
       0 \;\;& \text{if}\ c\in \set C^t\setminus\{\mathtt{b}\}\\
      {q}_x^{t}(i,c)/\sum_{k\in \set Y^{t-1}} q_x^{t}(i,k) \;\;& \text{if}\ c\in \set Y^{t-1}\,.
    \end{cases}
\end{equation}

The rationale behind $\ell_{kd}$ is that $f_{\theta^t}$ should produce activations close to the ones produced by $f_{\theta^{t-1}}$.
This regularizes the training procedure such that the parameters $\theta^t$ remain anchored to the solution found for classifying pixels of previous classes, \ie $\theta^{t-1}$.

The loss defined in Eq.~\eqref{eq:std-distill} has been used either in its base form or variants in different contexts, from incremental task \cite{li2017learning} and class learning \cite{rebuffi2017icarl,castro2018end} in object classification to complex scenarios such as detection \cite{shmelkov2017incremental} and segmentation \cite{michieli2019incremental}. Despite its success, it has a fundamental drawback in semantic segmentation: it completely ignores that {the representation of the background class evolves over time.} %
While with Eq.~\eqref{eq:our-CE} we tackled the first problem linked to the semantic shift of the background (\ie $\mathtt{b} \in \set T^{t}$ contains pixels of $\set Y^{t-1}$), we use the distillation loss to tackle the second: {annotations for background in $\set T^s$ with $s<t$ might include pixels of classes in $\set C^t$.}
{From the latter considerations, the background probabilities assigned to a pixel by the old predictor $f_{\theta^{t-1}}$ and by the current model $f_{\theta^{t}}$ do not share the same semantic content.}
More importantly, $f_{\theta^{t-1}}$ might predict as background pixels of classes in $\set C^t$ that we are currently trying to learn. Notice that this aspect is peculiar to the segmentation task and it is not considered in previous incremental learning models. 

{%
To address the semantic shift of the background class between the old and the current model, we explicitly revise the distillation loss in Eq.~\eqref{eq:std-distill}. In particular, we extend the reasoning behind Sec.\ref{sec:method-unknown} to the distillation soft-targets and we design a novel distillation loss by rewriting the probability distribution $\hat{p}_x^t(i,c)$ in Eq.~\eqref{eq:cases-kd} as:}
\begin{equation}
    \label{eq:cases-ukd}
    \hat{p}_x^{t}(i,c)= \begin{cases}
      {q}_x^{t}(i,c)\;\;& \text{if}\ c\neq\mathtt{b}\\
      \sum_{k\in \set C^t}q_x^{t}(i,k)\;\;& \text{if}\ c=\mathtt{b}\,.
    \end{cases}
\end{equation}
{We note that Eq.~\eqref{eq:cases-ukd} is identical to Eq.~\eqref{eq:cases-unknown} by setting $\set U = \set Y^t$ and by substituting the background class $b$ to $\mathtt{u}$.}

Similarly to Eq.~\eqref{eq:std-distill}, we still compare the probability of a pixel belonging to seen classes assigned by the old model, with its counterpart computed with the current parameters $\theta^t$. However, differently from classical distillation, in Eq.~\eqref{eq:cases-ukd} the probabilities obtained with the current model are kept unaltered, \ie normalized across the whole label space $\set {Y}^t$ and not with respect to the subset $\set Y^{t-1}$ (Eq.~\eqref{eq:cases-kd}). More importantly, the background class probability as given by $f_{\theta^{t-1}}$ is not directly compared with its counterpart in $f_{\theta^t}$, but with the probability of having \textit{either a new class or the background}, as predicted by $f_{\theta^t}$ ({see Fig. \ref{fig:method}, yellow block}).

We highlight that, with respect to Eq.~\eqref{eq:cases-kd} and other simple choices (\eg ignoring unlabeled pixels from Eq.~\eqref{eq:cases-kd}) this solution has two advantages. First, we can use the full output space of the old model to distill knowledge in the current one, without ignoring any pixel or class. Second, we can propagate the uncertainty we have on the semantic content of the background in $f_{\theta^{t-1}}$ without penalizing the probabilities of new classes we are learning in the current step $t$. 

\myparagraph{Classifiers' Parameters Initialization.} As discussed above, the background class $\con b$ is a special class devoted to collect the probability that a pixel belongs to an unknown object class. %
In practice, at each learning step $t$, the novel categories in $\set C^t$ are unknowns for the old classifier $f_{\theta^{t-1}}$. As a consequence, unless the appearance of a class in $\set C^t$ is very similar to one in $\set Y^{t-1}$, it is reasonable to assume that $f_{\theta^{t-1}}$ will likely assign pixels of $\set C^t$ to $\con b$. %
Taking into account this initial bias on the predictions of $f_{\theta^{t}}$ on pixels of $\set C^t$, it is detrimental to randomly initialize the classifiers for the novel classes. A random initialization would provoke a misalignment among the features extracted by the model (aligned with the background classifier) and the random parameters of the classifier itself. Notice that this could lead to possible training instabilities while learning novel classes since the network could initially assign high probabilities for pixels in $\set C^t$ to $\con b$.

To address this issue, we propose to initialize the classifier's parameters for the novel classes in such a way that given an image $x$ and a pixel $i$, the probability of the background $q_x^{t-1}(i,\con b)$ is uniformly spread among the classes in $\set C^{t}$, \ie $q_x^{t}(i,c)=q_x^{t-1}(i,\con b)/|\set C^t|\; \forall c \in \set C^t$, where $|\mathcal{C}^t|$ is the number of new classes (notice that $\mathtt{b} \in \mathcal{C}^t$). 
To this extent, let us consider a standard fully connected classifier and let us denote as $\{\omega^{t}_c, \beta^{t}_c\}\in\theta^t$ the classifier parameters for a class $c$ at learning step $t$, with $\omega$ and $\beta$ denoting its weights and bias respectively. We can initialize $\{\omega^t_c, \beta^t_c\}$ as follows:

\begin{align}
    \label{eq:init-cases}
    \omega_c^{t}&=\begin{cases}
     \omega_{\con b}^{t-1} \;\;& \text{if}\ c \in \mathcal{C}^t\\
      \omega^{t-1}_c \;\;& \text{otherwise}\\
    \end{cases}\\
        \label{eq:init-cases2}
    \beta_c^{t}&=\begin{cases}
      \beta_{\con b}^{t-1} - \log(|\set C^t|)\;\;& \text{if}\ c \in \mathcal{C}^t\\
      \beta^{t-1}_c \;\;& \text{otherwise}\\
    \end{cases}
\end{align}

where $\{\omega_\mathtt{b}^{t-1},\beta_\mathtt{b}^{t-1} \}$ are the weights and bias of the background classifier at the previous learning step. The fact that the initialization defined in Eq.\eqref{eq:init-cases} and \eqref{eq:init-cases2} leads to $q_x^{t}(i,c)=q_x^{t-1}(i,\con b)/|\set C^t|\; \forall c \in \set C^t$ is easy to obtain from $q_x^{t}(i,c)\propto \exp(\omega_\mathtt{b}^{t}\cdot x + \beta^{t}_\mathtt{b})$. 
 
As we will show in the experimental analysis, this simple initialization procedure brings benefits in terms of both improving the learning stability of the model and the final results, since it eases the role of the supervision imposed by Eq.\eqref{eq:our-CE} while learning new classes and follows the same principles used to derive our distillation loss (Eq.\eqref{eq:cases-ukd}).

\section{Semantic Segmentation using \comRev{Weak} Supervision}
\label{sec:weakly}
In the previous section, we revisited standard cross-entropy and distillation losses to take into account the prior we have on the content of the \comRev{unlabeled/background} pixels (for the cross-entropy loss) and the semantic of the predicted probabilities for the background class (for the distillation loss). The overall idea of the approach is that we can assume what is the set of semantic classes to which unlabeled/background pixels belong. This idea can be easily extended in other scenario where we can exploit partial annotations to impose priors on the content of unlabeled pixels. In the following, we will show how the same reasoning can be applied to tackle weakly supervised segmentation with point \comRev{and scribble} supervision.  %

\subsection{Problem Formulation}
{In weakly supervised segmentation using points \comRev{or scribbles} the goal is to obtain a model capable of predicting, for each pixel of the image, its correct semantic class, as in standard semantic segmentation. However, differently from the standard segmentation task, we train our model using a training set in which we do not have full pixel-level annotation, but just points \comRev{or scribbles}. In particular, for each instance of a class presented in a training image, only one \comRev{or few contiguous annotated pixels are} provided. Formally, considering an image $x$ and its label $y$ belonging to the training set $\set T$, the annotation is provided only for pixels $\set I^x_S = \{i : \;\forall\; i \in \set I\; \text{s.t.}\; y_i \in \set Y \}$, where $|\set I^x_S|<<|\set I|$. All the other image pixels are unlabeled. 

We address three weakly semantic segmentation setting: point-based \cite{bearman2016s} \comRev{and scribble-based \cite{lin2016scribblesup}} object segmentation, and point-based scene parsing \cite{qian2019weakly}.
The goal of object segmentation is to predict object classes in a target image, where the objects are countable things, such as \textit{cars}, \textit{bikes}, and \textit{dogs}. All the pixels that do not fall in these categories are labeled as background, which is considered a class in the output space $\mathcal{Y}$ of our model, similarly to Sec.~\ref{sec:incremental}. Formally, given a training set $\mathcal{T} \subset \mathcal{X} \times (\set Y \cup \con u)^N$, the goal is to learn a model able to predict, for each pixel $i$, a label $ y_i \in (\set Y)$. \comRev{Following the protocols defined in \cite{bearman2016s} and \cite{ lin2016scribblesup}, the point annotations are given only for the objects, while no points are provided for the background class. Differently, the scribble annotations also contain a scribble for the background class.} 

Scene parsing, instead, is a more complex task where the goal is to obtain a model able to predict both countable things and stuff classes (\ie all the non-countable classes, such as \textit{sky}, \textit{road}, \textit{ground}, etc.). In this setting, all the pixels in the image contain a semantic category and the background class is not included in the label space. Formally, the goal is to learn a model able to map each pixel $i$ to a label  $ y_i \in \set Y$. The mapping is learned using a training set $\mathcal{T} \subset \mathcal{X} \times (\set Y \cup \con u)^N$.
}

{
\subsection{Modeling the Unlabeled}
\label{sec:weakly-met}

Being provided few labeled pixels, previous approaches \cite{bearman2016s, qian2019weakly} proposed to apply a cross-entropy loss directly on the labeled points.
In particular, they defined a partial cross-entropy (PCE) loss that considered only the pixels for which an annotation is given. Formally, given an image $x$ and the respective annotation $y$, the PCE loss has the form:
\begin{equation} 
   \label{eq:pce}
   \ell_{PCE}(x, y) = - \frac{1}{|\set I^x_S|} \sum_{i \in \set I^x_S} \log {q}_x(i,y_i).
\end{equation}
This loss is crucial for the network to discriminate the classes and to localize them in the image. However, while this solution is simple and easy to implement, it completely discards the information provided by the unlabeled pixels. 
In Section \ref{sec:method-unknown} we showed a simple principle to extract value from them and in this section we will revisit the principle to adapt it in this scenario.

We start from the assumption that, for each instance of a class in the image it has been provided at least one labeled pixel. This assumption implies that we know which are the classes present in the image and that all the pixels in the image belong to one of those classes. Denoting the set of classes appearing in the label $y$ of an image $x$ as $\set U_x = \{c : \exists \,i\in I^x_S \; \text{s.t.}\; c = y_i \}$
, we can use the loss in Eq.~\eqref{eq:unknown}, with $\set U = \set U_x$ to consider the uncertainty we have on all the unlabeled pixels of the image. In particular, denoting as $\set I^x_u = \set I \setminus \set I^x_S$ the set of unlabeled pixels, we propose to extend Eq.~\eqref{eq:unknown} as follows:
\begin{equation} 
   \label{eq:unlabeled}
   \ell_{UNL}(x, y) = \ell_{PCE}(x,y) - \frac{\gamma}{|\set I^x_u|} \sum_{i \in \set I^x_u} \log p_x(i, u)\,,
\end{equation} 
where $p_x$ is the probability distribution defined in Eq.\eqref{eq:cases-unknown} with $\set U = \set U_x$, and $\gamma$ is a hyper-parameter to weight the importance of unlabeled pixels since in this scenario unlabeled pixels are many more than the labeled ones. %

Using the $\ell_{UNL}$ loss provides two important benefits to our training procedure when compared with  $\ell_{PCE}$: (i) information from labeled pixels is propagated to unlabeled ones, providing an additional source of supervision; (ii) if the network predicts an unlabeled pixel as belonging to a class $c$ which is not in the current image (\ie $c \notin \set U_x$), receives a feedback on the error from the loss function. %

To summarize, given a training set $\set T$, we train the network to minimize the following objective function:
\begin{equation}
  \label{eq:weakly-general}
    \mathcal{L}(\theta)= \frac{1}{|\mathcal{T}|} \sum_{(x,y)\in\mathcal{T}} \ell_{UNL}(x, y).
\end{equation}
We note that this loss function can be applied without any modification both \comRev{with point and scribble supervision to the object segmentation and to the scene parsing tasks}. The only difference relies on the classes contained in the image, since for object segmentation the background class $\con b$ belongs to every image, \ie $\con b \in \set U_x \;\; \forall x \in \set T$, while in scene parsing the background class is not considered.
As far as we know, our method is the first applicable on \comRev{point and scribble} supervision both on object segmentation and scene parsing, achieving state of the art results without relying on any other prior learned on additional data (e.g. objectness prior \cite{bearman2016s}). 
}

\section{Experiments}
\label{sec:experiments}
\subsection{{Datasets}}
{
In this work, we use three datasets: Pascal-VOC 2012, ADE20K and Cityscapes.
PASCAL-VOC 2012 \cite{pascal-voc-2012} is a widely used benchmark that includes 20 foreground object classes. We use the extra annotation provided in \cite{bharath2011sbd}, resulting in a dataset containing 10582 images in the training set and 1449 images in the validation.
ADE20K \cite{zhou2017scene} is a large-scale dataset that contains 150 classes. Differently from Pascal-VOC 2012, this dataset contains both stuff (\eg \textit{sky}, \textit{building}, \textit{wall}) and object classes. The dataset comprises more than 25K scene-centric images. Adopting the standard protocol \cite{zhao2017pyramid} we use 20K images for training and we reported the results on the 2K images of the validation set.
Cityscapes \cite{cordts2016cityscapes} is a dataset containing street-level images captured in central Europe that includes 19 classes, which are both objects or stuffs. The dataset provides high-resolution images with size $2048 \times 1024$, which are splitted in 2975 images for training, 500 for validation and 1525 for testing. However, since the test set ground truth are not available, we report results on the validation set as done by \cite{rota2018place}. We exclude from the training protocol the coarse-annotations, and we use only the fine-grained annotations.
}

\subsection{Incremental Learning in Semantic Segmentation}

\subsubsection{\icl\
 Baselines} \label{sec:baselines}
We compare our method against standard \icl\ baselines, 
 originally designed for classification tasks,
 on the considered segmentation task, 
 thus segmentation is treated as a pixel-level classification problem.
{Specifically, we report the results of six different regularization-based methods, three prior-focused and three data-focused approaches.} 

In the first category, we chose Elastic Weight Consolidation (EWC) \cite{kirkpatrick2017overcoming}, Path Integral (PI) \cite{zenke2017continual}, and Riemannian Walks (RW) \cite{chaudhry2018riemannian}. They employ different strategies to compute the importance of each parameter for old classes: EWC uses the empirical Fisher matrix, PI uses the learning trajectory, while RW combines EWC and PI in a unique model. We choose EWC since it is a standard baseline employed also in \cite{shmelkov2017incremental} and PI and RW since they are two simple applications of the same principle. Since these methods act at the parameter level, to adapt them to the segmentation task we keep the loss in the output space unaltered (\ie standard cross-entropy across the whole segmentation mask), computing the parameters' importance by considering their effect on learning old classes.

For the data-focused methods, we chose Learning without forgetting (LwF) \cite{li2017learning}, LwF multi-class (LwF-MC) \cite{rebuffi2017icarl} and the segmentation method of \cite{michieli2019incremental} (ILT).
We denote as LwF the original distillation based objective as implemented in Eq.\eqref{eq:obj-general} with basic cross-entropy and distillation losses, which is the same as \cite{li2017learning} except that distillation and cross-entropy share the same label space and classifier. LwF-MC is the single-head version of \cite{li2017learning} as adapted from \cite{rebuffi2017icarl}. It is based on multiple binary classifiers, with the target labels defined using the ground truth for novel classes (\ie $\set C^t$) and the probabilities given by the old model for the old ones (\ie $\set Y^{t-1}$). Since the background class is both in $\set C^t$ and $\set Y^{t-1}$ %
we implement LwF-MC by a weighted combination of two binary cross-entropy losses, on both the ground truth and the probabilities given by $f_{\theta^{t-1}}$. %
Finally, ILT \cite{michieli2019incremental} is the only method specifically proposed for \icl\ in %
segmentation. It uses a distillation loss in the output space, as in our adapted version of LwF \cite{li2017learning} and/or another distillation loss in the features space, attached to the output of the network decoder. Here, we use the variant where both losses are employed.
As done by \cite{shmelkov2017incremental}, we do not compare with replay-based methods (\eg \cite{rebuffi2017icarl}) since they violate the standard \icl\ assumption regarding the unavailability of old data.

In all tables we report other two baselines: simple fine-tuning (FT) on each $\set T^t$ (\eg Eq.\eqref{eq:CE}) and training on all classes offline (Joint). The latter can be regarded as an upper bound. In the tables we denote our method as \ours (\expandednick). All results are reported as mean Intersection-over-Union (mIoU) in percentage, averaged over all the classes of a learning step and all the steps.

\subsubsection{Implementation Details}\label{sec:impdetails}
For all methods we use the Deeplab-v3 architecture \cite{chen2017rethinking}. We use a ResNet-101 \cite{he2016deep} backbone for Pascal-VOC 2012 and ADE20K, and following \cite{rota2018place} a ResNeXt-101 \cite{xie2017aggregated} for Cityscapes. For both backbones we use an output stride of 16. Since memory requirements are an important issue in semantic segmentation, we use in-place activated batch normalization, as proposed in \cite{rota2018place}. The backbone has been initialized using the ImageNet pretrained model \cite{rota2018place}. %
We follow \cite{chen2017rethinking}, training the network with SGD and the same learning rate policy, momentum and weight decay. %
For ADE20K and Pascal-VOC 2012 we use an initial learning rate of $10^{-2}$ for the first learning step and $10^{-3}$ for the followings, as in \cite{shmelkov2017incremental}, while for Cityscapes we employed $2\times10^{-3}$ for the first step and $2\times10^{-4}$ in the following.
We train the model with a batch-size of 24 for 30 epochs for Pascal-VOC 2012, 60 epochs for ADE20K and 360 epochs for Cityscapes in every learning step. %
We apply the same data augmentation of \cite{chen2017rethinking} %
and we crop the images to $512\times 512$ during training. During test, we make a center crop $512\times 512$ of for Pascal-VOC 2012 and ADE20K, while we use the full-resolution images for Cityscapes.
For setting the \hypers of each method, we use the protocol of incremental learning defined in \cite{de2019continual}, using 20\% of the training set as validation.
The final results are reported on the standard validation set of the datasets.

\begin{table*}[t]
\centering
\setlength{\tabcolsep}{3pt} %
\small
\caption{Mean IoU (in \%) on the Pascal-VOC 2012 dataset for different incremental class learning scenarios.}
\label{tab:pascal}
\begin{tabular}{l||cc|c||cc|c||cc|c||cc|c||cc|c||cc|c}
\multicolumn{1}{c}{}    & \multicolumn{6}{c}{\textbf{{19-1}}}   & \multicolumn{6}{c}{{\textbf{15-5}}} & \multicolumn{6}{c}{{\textbf{15-1}}}    \\
\multicolumn{1}{c||}{}    & \multicolumn{3}{c||}{\bf{Disjoint}}        & \multicolumn{3}{c||}{\bf{Overlapped}}  & \multicolumn{3}{c||}{\bf{Disjoint}}     & \multicolumn{3}{c||}{\bf{Overlapped}}  & \multicolumn{3}{c||}{\textbf{Disjoint}}      & \multicolumn{3}{c}{\textbf{Overlapped}} \\
\bf{Method} & \it{1-19}  & \it{20}   & \it{all}  & \it{1-19}  & \it{20}   & \it{all}     & \it{1-15}  & \it{16-20}   & \it{all}   & \it{1-15}  & \it{16-20}   & \it{all} & \it{1-15}  & \it{16-20}   & \it{all}  & \it{1-15}  & \it{16-20}   & \it{all}     \\ \hline
{FT }     & 5.8     & 12.3    & 6.2     & 6.8       & 12.9      & 7.1      & 1.1    & 33.6    & 9.2      & 2.1     & 33.1  & 9.8    & 0.2        & 1.8        & 0.6        & 0.2        & 1.8        & 0.6       \\
{PI \cite{zenke2017continual}}     & 5.4     & 14.1    & 5.9     & 7.5       & 14.0      & 7.8      & 1.3    & 34.1    & 9.5      & 1.6     & 33.3  & 9.5    & 0.0 &	1.8 &	0.4   & 0.0        & 1.8        & 0.5 \\
{EWC \cite{kirkpatrick2017overcoming}}    & 23.2    & 16.0    & 22.9    & 26.9      & 14.0      & 26.3     & 26.7   & 37.7    & 29.4     & 24.3    & 35.5  & 27.1   & 0.3        & 4.3        & 1.3        & 0.3        & 4.3        & 1.3       \\
{RW \cite{chaudhry2018riemannian}}     & 19.4    & 15.7    & 19.2    & 23.3      & 14.2      & 22.9     & 17.9   & 36.9    & 22.7     & 16.6    & 34.9  & 21.2   & 0.2        & 5.4        & 1.5          & 0.0        & 5.2        & 1.3 \\
{LwF \cite{li2017learning} }    & 53.0    & 9.1     & 50.8    & 51.2      & 8.5       & 49.1     & 58.4   & 37.4    & 53.1     & 58.9    & 36.6  & 53.3   & 0.8        & 3.6        & 1.5        & 1.0        & 3.9        & 1.8       \\
{LwF-MC \cite{rebuffi2017icarl}} & 63.0    & 13.2    & 60.5    & 64.4      & 13.3      & 61.9     & 67.2   & 41.2    & 60.7     & 58.1    & 35.0  & 52.3   & 4.5        & 7.0        & 5.2    & 6.4        & 8.4        & 6.9 \\
{ILT \cite{michieli2019incremental}}    & 69.1    & 16.4    & 66.4    & 67.1      & 12.3      & 64.4     & 63.2   & 39.5    & 57.3     & 66.3    & 40.6  & 59.9   & 3.7        & 5.7        & 4.2        & 4.9        & 7.8        & 5.7       \\
\ours   & \bf{69.6} & \bf{25.6}   & \bf{67.4}      & \bf{70.2}    & \bf{22.1}       & \bf{67.8}    & \bf{71.8}     & \bf{43.3}    & \bf{64.7}    & \bf{75.5}     & \bf{49.4}  & \bf{69.0}  & \bf{46.2}       & \bf{12.9}       & \bf{37.9}       & \bf{35.1}       & \textbf{13.5}       & \textbf{29.7}             \\ \hline
{Joint} & 77.4 &	78.0&	77.4&	77.4& 78.0	& 77.4&	79.1&	72.6&	77.4&	79.1&	72.6&	77.4 & 79.1       & 72.6       & 77.4       & 79.1       & 72.6       & 77.4 \\

\end{tabular}
\end{table*}

\subsubsection{Pascal-VOC 2012} \label{sec:inc_pascal}
Following \cite{michieli2019incremental,shmelkov2017incremental}, we define two experimental settings, depending on how we sample images to build the incremental datasets.
Following \cite{michieli2019incremental}, we define an experimental protocol called the \textit{disjoint} setup: each learning step contains a unique set of images, whose pixels belong to classes seen either in the current or in the previous learning steps. Differently from \cite{michieli2019incremental}, at each step we assume to have only labels for pixels of novel classes, while the old ones are labeled as background in the ground truth. 
The second setup, that we denote as \textit{overlapped}, follows what done in \cite{shmelkov2017incremental} for detection: each training step contains all the images that have at least one pixel of a novel class, with only the latter annotated. It is important to note a difference with respect to the previous setup: images may now contain pixels of classes that we will learn in the future, but labeled as background. This is a more realistic setup since it does not make any assumption on the objects present in the images.

As done by previous works \cite{shmelkov2017incremental, michieli2019incremental}, we perform three different experiments concerning the addition of one class (\textit{19-1}), five classes all at once (\textit{15-5}), and five classes sequentially (\textit{15-1}), following the alphabetical order of the classes to split the content of each learning step. %

\myparagraph{Addition of one class \textit{(19-1)}.}
In this experiment, we perform two learning steps: the first in which we observe the first 19 classes, and the second where we learn the \textit{tv-monitor} class.
Results are reported in Table \ref{tab:pascal}. Without employing any regularization strategy, the performance on past classes drops significantly. FT, in fact, performs poorly, completely forgetting the first 19 classes. %
Unexpectedly, using PI as a regularization strategy does not provide benefits, while EWC and RW improve performance of nearly 15\%. However, prior-focused strategies are not competitive with data-focused ones. In fact, LwF, LwF-MC, and ILT, outperform them by a large margin, confirming the effectiveness of this approach on preventing catastrophic forgetting. While ILT surpasses standard \icl\ baselines, our model is able to obtain a further boost. This improvement is remarkable for new classes, where we gain $11\%$ in mIoU, while do not experience forgetting on old classes. It is especially interesting to compare our method with the baseline LwF which uses the same principles of ours but without modeling the background. Compared to LwF we achieve an average improvement of about $15\%$, thus demonstrating the importance of modeling the background in \icl\ for semantic segmentation. These results are consistent in both the \textit{disjoint} and \textit{overlapped} scenarios.%

\myparagraph{Single-step addition of five classes (\textit{15-5}).}
In this setting we add, after the first training set, the following classes: \textit{plant, sheep, sofa, train, tv-monitor}. Results are reported in Table \ref{tab:pascal}. 
Overall, the behavior on the first 15 classes is consistent with the 19-1 setting: FT and PI suffer a large performance drop, data-focused strategies (LwF, LwF-MC, ILT) outperform EWC and RW by far, while our method gets the best results, obtaining performances closer to the joint training upper bound. %
For what concerns the \textit{disjoint} scenario, our method improves over the best baseline of $4.6\%$ on old classes, of $2\%$ on novel ones and of $4\%$ in all classes. These gaps increase in the \textit{overlapped} setting where our method surpasses the baselines by nearly $10\%$ in all cases, clearly demonstrating its ability to take advantage of the information contained in the background class. %

\myparagraph{Multi-step addition of five classes (\textit{15-1}).}
This setting is similar to the previous one except that the last 5 classes are learned sequentially, one by one.
From Table \ref{tab:pascal} we can observe that performing multiple steps is challenging and existing methods work poorly for this setting, reaching performance inferior to 7\% on both old and new classes.
In particular, FT and prior-focused methods are unable to prevent forgetting, biasing their prediction completely towards new classes and demonstrating performances close to 0\% on the first 15 classes.
Even data-focused methods suffer a dramatic loss in performances in this setting, decreasing their score from the single to the multi-step scenarios of more than 50\% on all classes. %
On the other side, our method is still able to achieve good performances. %
Compared to the other approaches, \ours outperforms all baselines by a large margin in both old ($46.2\%$ on the \textit{disjoint} and $35.1\%$ on the \textit{overlapped}), and new (nearly $13\%$ on both setups) classes. As the overall performance drop ($11\%$ on all classes) shows, the \textit{overlapped} scenario is the most challenging one since it does not impose any constraint on which classes are present in the background. %
\begin{table}[t]
\small
\centering
\setlength{\tabcolsep}{2pt} %
\caption{Ablation study of the proposed method on the Pascal-VOC 2012 \textit{overlapped} setup. \textit{CE} and \textit{KD} denote our cross-entropy and distillation losses, while \textit{init} our initialization strategy.}%
\vspace{-5pt}
\label{tab:ablation}
\begin{tabular}{l||cc|c||cc|c||cc|c}
      \multicolumn{1}{c}{}& \multicolumn{3}{c}{\textbf{19-1}}          & \multicolumn{3}{c}{\textbf{15-5}}          & \multicolumn{3}{c}{\textbf{15-1}} \\
  &\it{1-19} & \it{20} & \it{all} & \it{1-15} & \it{16-20} & \it{all} & \it{1-15}  & \it{16-20}     & \it{all}     \\ \hline
LwF \cite{li2017learning}         & 51.2      & 8.5       & 49.1       & 58.9      & 36.6       & 53.3       & 1.0         & 3.9   & 1.8   \\
     + \textit{CE} & 57.6      & 9.9       & 55.2       & 63.2       & 38.1       & 57.0       & 12.0        & 3.7   & 9.9   \\
+ \textit{KD}&   66.0      & 11.9      & 63.3       & 72.9       & 46.3       & 66.3       & 34.8        & 4.5   & 27.2  \\
+ \textit{init} & \textbf{70.2}      & \textbf{22.1}      & \textbf{67.8}      & \textbf{75.5}      & \textbf{49.4}      & \textbf{69.0}      & \textbf{35.1}       & \textbf{13.5} & \textbf{29.7}
\end{tabular}
\end{table}

\begin{table*}[t]
\centering
\small
\setlength{\tabcolsep}{3pt} %
\caption{Mean IoU (in \%) on the ADE20K dataset for different incremental class learning scenarios.}
\label{tab:ade}
\begin{tabular}{l||cc|c||cccccc|c||ccc|c}
\multicolumn{1}{c}{} & \multicolumn{3}{c}{{\textbf{100-50}}} & \multicolumn{7}{c}{{\textbf{100-10}}} & \multicolumn{4}{c}{{\textbf{50-50}}} \\
\textbf{Method}       & \textit{1-100} & \textit{101-150} & \textit{all}  & \textit{1-100} & \textit{100-110} & \textit{110-120} & \textit{120-130} & \textit{130-140} & \textit{140-150} & \textit{all}  & \textit{1-50} & \textit{51-100} & \textit{101-150} & \textit{all}  \\ \hline
FT     & 0.0   & 24.9    & 8.3  & 0.0   & 0.0    & 0.0    & 0.0    & 0.0    & 16.6   & 1.1  & 0.0  & 0.0    & 22.0    & 7.3  \\
LwF \cite{li2017learning}   & 21.1  & 25.6    & 22.6 & 0.1   & 0.0    & 0.4    & 2.6    & 4.6    & 16.9   & 1.7  & 5.7  & 12.9   & 22.8    & 13.9 \\
LwF-MC \cite{rebuffi2017icarl} & 34.2  & 10.5    & 26.3 & 18.7  &	2.5 &	8.7 &	4.1 &	6.5 &	5.1 &	14.3  & 27.8 & 7.0    & 10.4    & 15.1 \\
ILT \cite{michieli2019incremental}  & 22.9  & 18.9    & 21.6 & 0.3   & 0.0    & 1.0    & 2.1    & 4.6    & 10.7   & 1.4  & 8.4  & 9.7    & 14.3    & 10.8 \\
MiB   & \textbf{37.9}  & \textbf{27.9}    & \textbf{34.6} & \textbf{31.8}  & \textbf{10.4}   & \textbf{14.8}   & \textbf{12.8}   & \textbf{13.6}   & \textbf{18.7}   & \textbf{25.9} & \textbf{35.5} & \textbf{22.2}   & \textbf{23.6}    & \textbf{27.0} \\ \hline
Joint  & 44.3  & 28.2    & 38.9 & 44.3  & 26.1   & 42.8   & 26.7   & 28.1   & 17.3   & 38.9 & 51.1 & 38.3   & 28.2    & 38.9 \\
\end{tabular}
\end{table*}

\myparagraph{Ablation Study.}
In Table \ref{tab:ablation} we report a detailed analysis of our contributions, considering the \textit{overlapped} setup. %
We start from the baseline LwF \cite{li2017learning} which employs standard cross-entropy and distillation losses.
We first add to the baseline our modified cross-entropy (\textit{CE}): this increases the ability to preserve old knowledge in all settings without harming (\textit{15-1}) or even improving (\textit{19-1}, \textit{15-5}) performances on the new classes. %
Second, we add our distillation loss (\textit{KD}) to the model. 
Our \textit{KD} provides a boost on the performances for both old and new classes. The improvement on old classes is remarkable, especially in the 15-1 scenario (\ie 22.8\%). For the novel classes, the improvement is constant and is especially pronounced in the 15-5 scenario (7\%). Notice that this aspect is peculiar of our \textit{KD} since standard formulation work only on preserving old knowledge. %
This shows that the two losses provide mutual benefits. %
Finally, we add our classifiers' initialization strategy (\textit{init}). This component provides an improvement in every setting, especially on novel classes: it doubles the performance on the \textit{19-1} setting ($22.1\%$ vs $11.9\%$) and triplicates on the \textit{15-1} ($4.5\%$ vs $13.5\%$). This confirms the importance of accounting for the background shift at the initialization stage to facilitate the learning of new classes. %

\subsubsection{ADE20K} \label{sec:inc_ade} %
We create the incremental datasets $\set T^t$ by splitting the whole dataset into disjoint image sets, without any constraint except ensuring a minimum number of images (\ie 50) where classes on $\set C^t$ have labeled pixels. %
Obviously, each $\set T^t$ provides annotations only for classes in $\set C^t$ while other classes (old or future) appear as background in the ground truth. %
In Table \ref{tab:ade} we report the mean IoU obtained averaging the results on two different class orders: the order proposed by \cite{zhou2017scene} and a random one. In this experiments, we compare our approach with data-focused methods only (\ie LwF, LwF-MC, and ILT) due to their gap in performance with prior-focused ones.

\begin{figure*}[t]
     \centering
     \begin{subfigure}[b]{0.96\textwidth}
         \centering
         \includegraphics[width=\textwidth]{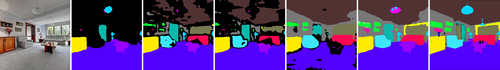}
     \end{subfigure}
     \hfill
     \begin{subfigure}[b]{0.96\textwidth}
         \centering
         \includegraphics[width=\textwidth]{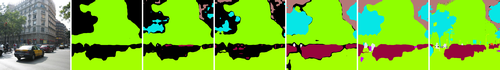}
     \end{subfigure}
     \hfill
     \begin{subfigure}[b]{0.96\textwidth}
         \centering
         \includegraphics[width=\textwidth]{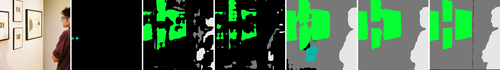}
     \end{subfigure}
          \hfill
     \begin{subfigure}[b]{0.96\textwidth}
         \centering
         \includegraphics[width=\textwidth]{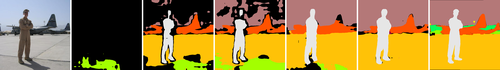}
     \end{subfigure}
        \vspace{-5pt}
     \caption{Qualitative results on the \textit{100-50} setting of the ADE20K dataset using different incremental methods. The image demonstrates the superiority of our approach on both new (\eg \textit{building}, \textit{floor}, \textit{table}) and old (\eg \textit{car}, \textit{wall}, \textit{person}) classes. From left to right: image, FT, LwF \cite{li2017learning}, ILT \cite{michieli2019incremental}, LwF-MC \cite{rebuffi2017icarl}, our method, and the ground-truth. Best viewed in color.}
        \label{fig:qualitative}
    \vspace{-10pt}
\end{figure*}

\myparagraph{Single-step addition of 50 classes (\textit{100-50}).} In the first experiment, we initially train the network on 100 classes and we add the remaining 50 all at once.
From Table \ref{tab:ade} we can observe that FT is clearly a bad strategy on large scale settings since it completely forgets old knowledge. %
Using a distillation strategy enables the network to reduce the catastrophic forgetting: LwF obtains $21.1\%$ on past classes, ILT $22.9\%$, and LwF-MC $34.2\%$. Regarding new classes, LwF is the best strategy, exceeding LwF-MC by $18.9\%$ and ILT by $6.6\%$.
However, our method is far superior to all others, improving on the first classes and on the new ones. Moreover, we can observe that we are close to the joint training upper bound, especially considering new classes, where the gap with respect to it is only $0.3\%$.
In Figure \ref{fig:qualitative} we report some qualitative results which demonstrate the superiority of our method compared to the baselines.

\myparagraph{Multi-step addition of 50 classes (\textit{100-10}).} We then evaluate the performance on multiple incremental steps: we start from 100 classes and we add the remaining classes 10 by 10, resulting in 5 incremental steps. In Table \ref{tab:ade} we report the results on all sets of classes after the last learning step.
In this setting the performance of FT, LwF and ILT are very poor because they strongly suffers catastrophic forgetting. %
LwF-MC demonstrates a better ability to preserve knowledge on old classes, at the cost of a performance drop on new classes.
Again, our method achieves the best trade-off between learning new classes and preserving past knowledge, outperforming LwF-MC by $11.6\%$ considering all classes. %

\myparagraph{Three steps of 50 classes (\textit{50-50}).} Finally, in Table \ref{tab:ade} we analyze the performance on three sequential steps of 50 classes.
Previous \icl\ methods achieve different trade-offs between learning new classes and not forgetting old ones. LwF and ILT obtain a good score on new classes, but they forget old knowledge. On the contrary, LwF-MC preserves knowledge on the first 50 classes without being able to learn new ones.
Our method outperforms all the baselines by a large margin with a gap of $11.9\%$ on the best performing baseline, achieving the highest mIoU on every step. Remarkably, the highest gap 
is on the intermediate step, where there are classes that we must %
 both learn incrementally and preserve from forgetting on the subsequent learning step. %

{
\begin{table*}[t]
\centering
\small
\caption{Mean IoU (in \%) on the Cityscapes dataset for different incremental class learning scenarios.}
\label{tab:cts}
\begin{tabular}{l||cc|c||cc|c||ccccc|c}
\multicolumn{1}{c}{} & \multicolumn{3}{c}{{\textbf{vehicles}}} & \multicolumn{3}{c}{{\textbf{non-driving}}} & \multicolumn{6}{c}{\textbf{11-2}} \\
\textbf{Method}       & \textit{old} & \textit{novel} & \textit{all}  & \textit{old} & \textit{novel} & \textit{all}  & \textit{1-11} & \textit{12-13} & \textit{14-15} & \textit{16-17} & \textit{18-19} & \textit{all}  \\ \hline
FT 	    & 0.0  & \textbf{71.0} &	22.4 &	0.0	 &  69.0 &	21.8 &	 0.0 &	 0.0 &	 0.0 &	 0.0 &	\textbf{57.3} &	6.0  \\
LwF \cite{li2017learning}     & 69.0 & 44.4 &	61.3 &	63.9 &	63.1 &	63.6 &	27.8 &	 0.0 &	 4.8 &	38.5 &	49.7 &	25.9 \\
LwF-MC \cite{rebuffi2017icarl}	&  58.9	& 47.0 &	55.2 &	48.7 &	58.5 &	51.8&	60.6 &	 0.0 &	 0.0 &	 9.6 &	33.5 &	39.6 \\
ILT \cite{michieli2019incremental}	    & 68.3 & 37.4 &	58.5 &	64.9 &	54.8 &	61.7 &	28.9 &	 0.0 &	 6.8 &	27.3 &	33.2 &	23.8 \\
MiB	    & \textbf{69.4} & 66.6 &	\textbf{68.5} &	\textbf{66.4} &	\textbf{70.0} &	\textbf{67.6} &	\textbf{70.2} &	\textbf{33.7} &	\textbf{53.7} & \textbf{49.0} &	53.9 &	\textbf{60.7} \\ \hline
Joint	 & 72.8 & 73.8 &	73.1 &	72.5 &	74.3 &	73.1 &	73.6 &	68.3 &	79.0 &	72.4 &	69.9 &	73.1 \\ 
\end{tabular}
\end{table*}
\subsubsection{Cityscapes} \label{sec:inc_cityscapes}
As done for the ADE20K dataset, we split the dataset into disjoint sets, one for each learning step $t$. Annotations are provided only for classes in $\set C^t$ while other classes (old or future) appear as background in the ground truth.
Also for Cityscapes, we compare our method with data-focused methods only (\ie LwF, LwF-MC, and ILT).
Table \ref{tab:cts} reports the mean IoU obtained on three different settings: \textit{vehicles}, \textit{non-driving}, and \textit{11-2}.

\myparagraph{Addition of vehicles classes (\textit{vehicles}).}
In the first setting, we initially train the network on the non-vehicles classes of Cityscapes (\ie \textit{road, sidewalk, building, wall, fence, pole, light, sign, vegetation, terrain, sky, person, rider}) and then we add in a single step all the vehicle classes (\ie \textit{car, truck, bus, train, motorcycle, bicycle}).
From Table \ref{tab:cts}, we note that fine-tuning the network on the novel classes gives good results, but at cost of completely forgetting the old classes. Adding a distillation strategy to it improves the results, especially on old classes. In particular, LwF and ILT obtain respectively $69.0$\% and $68.3$\%, while LwF-MC achieves a lower mIoU $58.9$\% but it is the highest among the three on the novel classes, achieving $47.0$\%. 
Comparing MiB with the other methods, it is able to maintain a good performance on old classes, achieving the best result $69.4$\%, while it is also able to learn properly the novel classes, being inferior to FT only of $4.4$\%. Overall, the best method is MiB, exceeding other methods more than $7.2$\% mIoU on all classes and being inferior to the joint training upper-bound only by $4.6$\%.

\myparagraph{Addition of non-driving classes (\textit{non-driving}).}
In this experiment we use the same number of incremental classes as the previous but we propose a different grouping. The classes are semantically divided in two groups, depending if they are strictly related to driving or not. The first group, which we train first, is made by \textit{driving} classes: \textit{road, sidewalk, pole, light, sign, person, rider, car, truck, bus, train, motorcycle and bicycle}. The second group, which is learned incrementally, contains \textit{non-driving} classes: \textit{building, wall, fence, vegetation, terrain, sky}.
As can be noted in Table \ref{tab:cts}, the results are coherent with the findings on the previous setting. FT performs well on novel classes, while it completely forgets about old ones. LwF, LwF-MC, and ILT achieve good performances on both old and novel classes. In particular, the best among the three is LwF, which obtains $63.9$\% on the \textit{driving} classes, $63.1$\% on the \textit{non-driving} classes and an overall mIoU of $63.6$\%.
However, the best method is MiB, which exceeds LwF both on \textit{driving} ($2.5$\%) and \textit{non-driving} ($6.9$\%) classes. Overall, it achieves $67.6$\% mIoU, which is inferior to the upper-bound of $5.5$\%. 

\myparagraph{Multi-step addition of 2 classes (\textit{11-2}).}
Finally, we analyze the performance on a multi-step setting, where we add two classes in four different steps. We start from 11 classes (\textit{road, sidewalk, building, wall, fence, pole, light, sign, vegetation, terrain, sky}) and then we add \textit{person and rider}, then \textit{car and truck}, then \textit{bus and train}, and finally \textit{motorcycle and bicycle}.
In Table \ref{tab:cts}, we report the results for each group of classes, after all the classes have been learned. As before, fine-tuning the network provide good performance on the novel classes but it suffers catastrophic forgetting and obtains $0.0$\% mIoU on old classes. LwF-MC obtains a good results on the first set of classes (\textit{1-11}) but it struggles to learn the novel ones, especially considering the intermediate classes. LwF and ILT demonstrate a similar behavior, forgetting old classes both on the first and intermediate steps. However, LwF achieves better results on the novel ones, exceeding ILT by $16.5$\%.
Our method outperforms all the other baselines by more than $21$\% mIoU. In particular, it is the only method the only method able to maintain good performances on the intermediate steps.
}

\subsection{Semantic Segmentation with \comRev{Weak} Supervision}

\begin{table}[t]
\small
\centering
\caption{Results on point-based weakly supervised object segmentation on Pascal-VOC (mIoU in \%).} 
\label{tab:weakly-voc}
\begin{tabular}{l|cc}
\textbf{Method} & \textbf{mIoU} & \textbf{P-Acc}  \\ \hline
Img Lvl	\cite{bearman2016s}             & 33.2 & 76.0 \\
Img Lvl + PCE \cite{bearman2016s}	    & 34.7 & 58.9 \\
Img Lvl + PCE + Obj \cite{bearman2016s} & 42.1 & 81.5 \\ \hline
PCE + bkg                               & 38.8 & 81.9 \\ \hline
MiB (lr $10^{-5}$)	                    & 45.3 & 82.3 \\
MiB (lr $10^{-4}$)	                    & \textbf{46.7} & \textbf{83.6} \\ \hline
Full Supervision	                    & 58.8 & 89.9 \\
\end{tabular}
\end{table}

\begin{table}[t]
\small
\centering
\caption{\comRev{Results on scribble-based weakly supervised object segmentation on Pascal-VOC (mIoU in \%).}}
\label{tab:scribble}
{
\begin{tabular}{l|cc}
\textbf{Method} & \textbf{wo/ CRF} & \textbf{w/ CRF}  \\  \hline 
PCE  & 69.5 & 72.8 \\ 
MiB  & \textbf{72.3} & \textbf{75.1} \\\hline
Scribble-Sup \cite{lin2016scribblesup} & - & 63.1 \\
NormalizedCut \cite{tang2018normalized}  & 72.8 & 74.5 \\
KernelCut \cite{tang2018regularized} & 73.0 & 75.0 \\
BPG \cite{wang2019boundary}  & \textbf{73.2} & \textbf{76.0} \\ \hline
Full Supervision & 75.8 & 76.4 \\

\end{tabular}}
\end{table}

\begin{table}[t]
\small
\centering
\caption{Results on point-based weakly supervised scene parsing on ADE20K (mIoU in \%).} 
\label{tab:weakly-ade}
\begin{tabular}{l|cc||cc}
\multicolumn{1}{c}{} & \multicolumn{2}{c}{Our protocol} & \multicolumn{2}{c}{\cite{qian2019weakly} protocol} \\
\textbf{Method} & \textbf{mIoU} & \textbf{P-Acc} & \textbf{mIoU} & \textbf{P-Acc} \\ \hline
PCE	                                    & 22.4 & 60.9 & 20.2 (17.7) & 58.3 (58.0) \\
PDML \cite{qian2019weakly}       	    & 21.1 & 56.6 & 19.3 (19.6) & 55.5 (61.0) \\ \hline
MiB         	                        & \textbf{22.9} & \textbf{62.2} & \textbf{21.0} & \textbf{59.5} \\ \hline
Full Supervision	                    & 29.7 & 68.8 & 25.1 & 66.0 \\

\end{tabular}
\end{table}

\subsubsection{Point-based Object Segmentation on Pascal-VOC} \label{sec:ws_objsec}
Following the work of \cite{bearman2016s}, we evaluate our method on object segmentation using the Pascal-VOC dataset and the point annotations the authors' provide. 
Differently from \cite{bearman2016s}, we employ a Resnet-101 \cite{he2016deep} as our backbone, with the modification of dilated convolutions, as in standard state-of-the-art architectures \cite{chen2017rethinking}. To recover the input resolution, we add after a bilinear interpolation layer on top of the Resnet-101, without additional trainable parameters.
We initialize the backbone with an ImageNet pretrained model, as in \cite{bearman2016s}, using the weights provided by \cite{rota2018place}. However, differently from them, we did not initialize the classifier since we were not able to establish the correct mapping among the ImageNet indices published by \cite{bearman2016s} and the ImageNet classes.
For fairness of comparison, we implemented \cite{bearman2016s} using our same backbone and training protocol.
We follow \cite{chen2017rethinking} and we train the network using SGD with momentum $0.9$, weight decay $10^{-4}$, and the same learning rate polynomial policy $base\_lr \times (1- \frac{iteration}{max\_iterations})^{0.9}$. We use an initial learning rate of $10^{-5}$ for the methods in \cite{bearman2016s} and $10^{-4}$ for the fully-supervised baseline. We report the results for our method with both learning rates. 
For all the methods we train the network using a batch size of $24$ for $30$ epochs. We crop the images to $512 \times 512$ during training and we apply the same data augmentation of \cite{chen2017rethinking}.

\myparagraph{Results.} \label{sec:weakly-voc-results}
In Table \ref{tab:weakly-voc} we report the mIoU and the overall pixel accuracy (\textit{P-Acc}).
The first three rows of the table refer to the methods proposed in \cite{bearman2016s}. In particular, we refer to \textit{Img Lvl} as the model trained only using Eq.2 of \cite{bearman2016s} which does not consider points location, but only image-level labels. This method achieves $33.2$\% mIoU, which is $4.4$\% better than the one reported by \cite{bearman2016s}. In the second row, we add the partial cross-entropy (PCE) loss, as proposed by \cite{bearman2016s}, and we refer to it as \textit{Img Lvl + PCE}. For this method, we use all the points available in the annotation and we do not weight them ($\alpha_i=1, \forall i \in \set I_S$). Adding the PCE improves the mIoU of $1.5$\% but deteriorates the pixel accuracy by $17.1$\%. This is due to the bias of the model toward the semantic classes, which led the model to assign an object label even to background pixels, which are the majority. However, introducing the Objectness Prior (\textit{Img Lvl + PCE + Obj}) \cite{bearman2016s} computed on an additional dataset (following \cite{bearman2016s}) improves the results, achieving $42.1$\% mIoU and $81.5$\% pixel accuracy. %

Nonetheless, our method outperforms all the three variants of \cite{bearman2016s}. In particular, we report it twice to be fair in the comparison: \textit{MiB (lr $10^{-5}$)} employs the same learning rate of \cite{bearman2016s}, while \textit{MiB (lr $10^{-4}$)} uses a learning rate $10^{-4}$ which we found better %
With both learning rates, MiB achieves better performance than \cite{bearman2016s}, demonstrating that our method is better in modeling the unknown pixels. In particular, \textit{MiB (lr $10^{-4}$)} achieves $46.7$\% mIoU and $83.6$\% pixel accuracy, being inferior to the fully supervised baselines of $12.1$\% $6.3$\% respectively. We would like to highlight that, differently from \cite{bearman2016s}, MiB does not use any objectness prior.

Finally, to prove that the improvement of our method is given by the way we model unlabeled pixels and not by rescaling the contribution of the background, we introduce the baseline referred as \textit{PCE + bkg}. In this method, we still use Eq.~\eqref{eq:weakly-general}, but we consider as possible class for the unlabeled pixels only the background. %
As can be noted in Table \ref{tab:weakly-voc}, this method is not able to learn properly the classes, obtaining $38.8$\% mIoU, which is $7.9$\% less than \textit{MiB (lr $10^{-4}$)}. %
\addnote[res:pcebkg]{1}{In particular, considering all the unlabeled pixels as background biases PCE+bkg toward this class. %
Instead, MiB models the unlabeled pixels using the prior given by the point labels, \ie $\set{U}=\set{U}_x$, pushing the network to predict them either as background or as any of the annotated classes.
}

\subsubsection{\comRev{Scribble-based Object Segmentation on Pascal-VOC}} \label{sec:exp_scribble}
\addnote[exp:scribble]{1}{
To evaluate our method on scribble-supervised semantic segmentation we followed the experimental protocol defined in \cite{tang2018regularized, wang2019boundary}, using the Pascal-VOC 2012 dataset and the scribble annotation released by \cite{lin2016scribblesup}. We employ the Deeplab-v2 architecture \cite{chen2017deeplab} with the Resnet-101 backbone \cite{he2016deep}. As in \cite{tang2018regularized, wang2019boundary}, we use dilated convolutions obtaining an output resolution 8 times smaller than the input. Moreover, we follow the strategy used in \cite{wang2019boundary} and we train the network on a single-scale resolution using a polynomial learning rate policy $base\_lr \times (1- \frac{iteration}{max\_iterations})^{0.9}$ with a batch size of 10 images and with $base\_lr=2.5\times10^{-4}$, momentum 0.9 and weight decay $5\times10^{-4}$. We train the network for 20K iterations using $321\times321$ cropped images, after applying horizontally flip (left-right) and randomly scaling the input images (from 0.5 to 2.0). In the testing stage, similarly to previous works \cite{tang2018regularized, wang2019boundary} we use multi-scale inputs (\ie [0.5, 0.75, 1.0, 1.25, 1.5]) with max voting to get the final prediction.}

\myparagraph{\comRev{Results}} \label{sec:weakly-scribble-results}
\addnote[exp:scribble-res]{1}{
Table \ref{sec:weakly-scribble-results} reports the mIoU with and without applying the dense CRF \cite{krahenbuhl2011efficient} post-processing using scribble-supervision. The top part reports the results of methods not explicitly designed for the scribble annotation (i.e. the PCE baseline and MiB), while the following reports the scribble-specific state-of-the-art approaches \cite{lin2016scribblesup, tang2018normalized, tang2018normalized, wang2019boundary}, and the fully-supervised upper-bound.
As for point supervision, the PCE baseline trains the network using the cross-entropy only on labeled pixels, as described in Eq.\eqref{eq:pce}. We note that PCE is already a competitive baseline, obtaining 72.8\% mIoU, \ie 3.6\% below the fully-supervised upper bound (76.4\%), demonstrating that the model is able to extract meaningful information even from few pixels. 
However, introducing our loss as reported in Eq.~\eqref{eq:weakly-general}, we are able to outperform the PCE baseline. In particular, MiB obtains 72.3\% (+1.8\% \wrt PCE) without CRF and 75.1\% (+2.3) with CRF. This remarks that unlabeled pixels bring crucial information %
to improve the results.  
\newline \indent Comparing MiB with the state-of-the-art methods, we note that it achieves competitive performance. In particular, comparing with NormalizedCut \cite{tang2018normalized} and KernelCut \cite{tang2018regularized}, we see that MiB obtains inferior performance without using the CRF, but it achieves superior performance while using it (+0.6\% \wrt NormalizedCut and +0.1\% \wrt KernelCut). We argue that KernelCut and NormalizedCut are superior to MiB without CRF since they already integrate the CRF in their training objective to better model the boundaries. However, the CRF post-processing is useful to correct boundary predictions and improves MiB performance while having less impact on NormalizedCut and KernelCut.
Finally, BPG \cite{wang2019boundary} achieves better results than MiB both without (+0.9\%) and with (+0.9\%) CRF post-processing. However, we remark that BPG introduces two sub-networks in the segmentation architecture to model boundaries, largely increasing the number of parameters and requiring additional supervision for boundary prediction. On the contrary, MiB is a general method that introduces only a loss function on unlabeled pixels, without requiring either to modify the network architecture or additional supervision.}

\subsubsection{Scene Parsing on ADE20K}
We evaluate our method also on the scene parsing task, as proposed by \cite{qian2019weakly}. The task is based on the ADE20K dataset and on the point annotation used by \cite{qian2019weakly}, which have been released in the LID Challenge 2020 \footnote{https://lidchallenge.github.io/challenge.html, see track 2.}.
Since the code of \cite{qian2019weakly} has not been released, we re-implemented it following the details and the algorithm provided in the paper.
Moreover, we report the results using two different training protocols, since we noted that the protocol of \cite{qian2019weakly} was sub-optimal.
Both protocols employ a Resnet-101 \cite{he2016deep} architecture with dilated convolutions, followed by a bilinear interpolation layer to recover the input resolution, as proposed by \cite{qian2019weakly}. 
The first protocol we implemented is the same of \cite{qian2019weakly}. The network is trained using SGD with momentum $0.9$, weight decay $5\times10^{-4}$ and an initial learning rate of $2.5 \times 10^{-4}$ that is decayed following a polynomial schedule $base\_lr \times (1- \frac{iteration}{max\_iterations})^{0.8}$. The dataset is iterated using a batch size of $16$ and the images are randomly cropped with size $321 \times 321$. However, the number of epochs has not been specified in \cite{qian2019weakly} and we train the network for 60 epochs.
The second protocol follows the protocol and hyperparameters described in Sec. \ref{sec:ws_objsec}. We only change the base learning rate that we set to $10^{-3}$.

\myparagraph{Results.}
The results are shown in Table \ref{tab:weakly-ade} where we report the mean Intersection-over-Union (mIoU) and the overall pixel accuracy (P-Acc). In the bracket we reported the numbers as reported in \cite{qian2019weakly}.
Following \cite{qian2019weakly}, we implemented the partial cross entropy (PCE) baseline, which only applies the cross-entropy loss on the pixels to whom a label is provided, as described in Eq.\ref{eq:pce}. As observed by \cite{qian2019weakly}, this is a strong baseline for point-supervised methods: it achieves $22.4$\% mIoU using our protocol and $20.2$\% mIoU on the one of \cite{qian2019weakly}. However, we note that the result obtained by this baseline are better than the one found in \cite{qian2019weakly} with a gap of $2.5$\% mIoU and $0.3$\% pixel accuracy.  
The PDML \cite{qian2019weakly} baseline obtains results in line with \cite{qian2019weakly}. However, comparing it with the PCE baseline, it perform worse, exhibiting a drop of performance of $1.3$\% and $0.9$\% mIoU in the two protocols. 
However, our method outperforms both baselines. It achieves $22.9$\% mIoU using our protocol, which is $0.5$\% more than PCE, and $21.0$\% using the \cite{qian2019weakly} protocol, with a gap of $0.8$\% with respect to PCE.

\section{Conclusions}
\label{sec:conclusions}
In this work, we proposed a general loss function for semantic segmentation under partial or weak supervision. This formulation considers unlabeled pixels as ground-truth annotation for \textit{any} possible class that pixel might contain. We considered two application scenarios for the method, incremental class learning and point-based weakly supervised semantic segmentation. In incremental class learning, we %
analyze the realistic scenario where the new training set does not provide annotations for old classes, leading to the semantic shift of the background class and exacerbating the catastrophic forgetting problem.
We addressed this issue by 
revisiting standard distillation-based \icl\ algorithms with our general principle in both cross-entropy and distillation losses, where the uncertainty on the unlabeled/background pixels is on the presence of old classes for the former and of new classes for the latter. Additionally, we propose a classifiers' initialization strategy which allows our network to explicitly model the semantic shift of the background. %
Results show that
our approach outperforms regularization-based \icl\ methods by a large margin, considering both small and large scale datasets. 

In a second series of experiments, we apply our general formulation to semantic segmentation with point \comRev{and scribble} supervision, where the prior on unlabeled pixels is given by the set of classes present in the current image. We show how our model \comRev{obtains competitive performance with respect} previous approaches in both objects segmentation and scene parsing, without any additional prior on the objects \comRev{or without making assumptions on the provided annotations.}

Future works might consider the application of the approach under different levels of weak supervision (e.g. bounding boxes \cite{khoreva2017simple}, polygons \cite{cordts2016cityscapes}) and on new tasks with partial knowledge on the unlabeled pixels, such as zero-shot learning \cite{xian2019semantic}.

\ifCLASSOPTIONcompsoc
  \section*{Acknowledgments}
\else
  \section*{Acknowledgment}
\fi

We acknowledge financial support from ERC grant 637076 - RoboExNovo obtained by Barbara Caputo. This work has been partially funded by the ERC (853489-DEXIM) and the DFG (2064/1–Project number 390727645). %
Computational resources were partially provided by HPC@POLITO\footnote{http://www.hpc.polito.it}.

\ifCLASSOPTIONcaptionsoff
  \newpage
\fi
 
\bibliographystyle{IEEEtran}
\bibliography{main.bib}

\begin{thebibliography}{10}
\providecommand{\url}[1]{#1}
\csname url@samestyle\endcsname
\providecommand{\newblock}{\relax}
\providecommand{\bibinfo}[2]{#2}
\providecommand{\BIBentrySTDinterwordspacing}{\spaceskip=0pt\relax}
\providecommand{\BIBentryALTinterwordstretchfactor}{4}
\providecommand{\BIBentryALTinterwordspacing}{\spaceskip=\fontdimen2\font plus
\BIBentryALTinterwordstretchfactor\fontdimen3\font minus
  \fontdimen4\font\relax}
\providecommand{\BIBforeignlanguage}[2]{{%
\expandafter\ifx\csname l@#1\endcsname\relax
\typeout{** WARNING: IEEEtran.bst: No hyphenation pattern has been}%
\typeout{** loaded for the language `#1'. Using the pattern for}%
\typeout{** the default language instead.}%
\else
\language=\csname l@#1\endcsname
\fi
#2}}
\providecommand{\BIBdecl}{\relax}
\BIBdecl

\bibitem{long2015fully}
J.~Long, E.~Shelhamer, and T.~Darrell, ``Fully convolutional networks for
  semantic segmentation,'' in \emph{CVPR}, 2015.

\bibitem{pascal-voc-2012}
M.~Everingham, L.~Van~Gool, C.~K.~I. Williams, J.~Winn, and A.~Zisserman, ``The
  {PASCAL} {V}isual {O}bject {C}lasses {C}hallenge 2012 {(VOC2012)}
  {R}esults,''
  http://www.pascal-network.org/challenges/VOC/voc2012/workshop/index.html.

\bibitem{zhou2017scene}
B.~Zhou, H.~Zhao, X.~Puig, S.~Fidler, A.~Barriuso, and A.~Torralba, ``Scene
  parsing through ade20k dataset,'' in \emph{CVPR}, 2017.

\bibitem{chen2018encoder}
L.-C. Chen, Y.~Zhu, G.~Papandreou, F.~Schroff, and H.~Adam, ``Encoder-decoder
  with atrous separable convolution for semantic image segmentation,'' in
  \emph{ECCV}, 2018.

\bibitem{zhao2017pyramid}
H.~Zhao, J.~Shi, X.~Qi, X.~Wang, and J.~Jia, ``Pyramid scene parsing network,''
  in \emph{CVPR}, 2017.

\bibitem{lin2017refinenet}
G.~Lin, A.~Milan, C.~Shen, and I.~Reid, ``Refinenet: Multi-path refinement
  networks for high-resolution semantic segmentation,'' in \emph{CVPR}, 2017.

\bibitem{zhang2018exfuse}
Z.~Zhang, X.~Zhang, C.~Peng, X.~Xue, and J.~Sun, ``Exfuse: Enhancing feature
  fusion for semantic segmentation,'' in \emph{ECCV}, 2018.

\bibitem{chen2017rethinking}
L.-C. Chen, G.~Papandreou, F.~Schroff, and H.~Adam, ``Rethinking atrous
  convolution for semantic image segmentation,'' 2017.

\bibitem{chen2017deeplab}
L.-C. Chen, G.~Papandreou, I.~Kokkinos, K.~Murphy, and A.~L. Yuille, ``Deeplab:
  Semantic image segmentation with deep convolutional nets, atrous convolution,
  and fully connected crfs,'' \emph{IEEE T-PAMI}, vol.~40, no.~4, pp. 834--848,
  2017.

\bibitem{chen2016attention}
L.-C. Chen, Y.~Yang, J.~Wang, W.~Xu, and A.~L. Yuille, ``Attention to scale:
  Scale-aware semantic image segmentation,'' in \emph{CVPR}, 2016.

\bibitem{kolesnikov2016seed}
A.~Kolesnikov and C.~H. Lampert, ``Seed, expand and constrain: Three principles
  for weakly-supervised image segmentation,'' in \emph{European conference on
  computer vision}.\hskip 1em plus 0.5em minus 0.4em\relax Springer, 2016, pp.
  695--711.

\bibitem{bearman2016s}
A.~Bearman, O.~Russakovsky, V.~Ferrari, and L.~Fei-Fei, ``What’s the point:
  Semantic segmentation with point supervision,'' in \emph{European conference
  on computer vision}.\hskip 1em plus 0.5em minus 0.4em\relax Springer, 2016,
  pp. 549--565.

\bibitem{rebuffi2017icarl}
S.-A. Rebuffi, A.~Kolesnikov, G.~Sperl, and C.~H. Lampert, ``icarl: Incremental
  classifier and representation learning,'' in \emph{CVPR}, 2017.

\bibitem{mccloskey1989catastrophic}
M.~McCloskey and N.~J. Cohen, ``Catastrophic interference in connectionist
  networks: The sequential learning problem,'' in \emph{Psychology of learning
  and motivation}.\hskip 1em plus 0.5em minus 0.4em\relax Elsevier, 1989,
  vol.~24, pp. 109--165.

\bibitem{li2017learning}
Z.~Li and D.~Hoiem, ``Learning without forgetting,'' \emph{IEEE T-PAMI},
  vol.~40, no.~12, pp. 2935--2947, 2017.

\bibitem{castro2018end}
F.~M. Castro, M.~J. Mar{\'\i}n-Jim{\'e}nez, N.~Guil, C.~Schmid, and K.~Alahari,
  ``End-to-end incremental learning,'' in \emph{ECCV}, 2018.

\bibitem{hinton2015distilling}
G.~Hinton, O.~Vinyals, and J.~Dean, ``Distilling the knowledge in a neural
  network,'' 2015.

\bibitem{cordts2016cityscapes}
M.~Cordts, M.~Omran, S.~Ramos, T.~Rehfeld, M.~Enzweiler, R.~Benenson,
  U.~Franke, S.~Roth, and B.~Schiele, ``The cityscapes dataset for semantic
  urban scene understanding,'' in \emph{Proceedings of the IEEE conference on
  computer vision and pattern recognition}, 2016, pp. 3213--3223.

\bibitem{lin2016scribblesup}
D.~Lin, J.~Dai, J.~Jia, K.~He, and J.~Sun, ``Scribblesup: Scribble-supervised
  convolutional networks for semantic segmentation,'' in \emph{Proceedings of
  the IEEE Conference on Computer Vision and Pattern Recognition}, 2016, pp.
  3159--3167.

\bibitem{qian2019weakly}
R.~Qian, Y.~Wei, H.~Shi, J.~Li, J.~Liu, and T.~Huang, ``Weakly supervised scene
  parsing with point-based distance metric learning,'' in \emph{Proceedings of
  the AAAI Conference on Artificial Intelligence}, vol.~33, 2019, pp.
  8843--8850.

\bibitem{cermelli2020modeling}
F.~Cermelli, M.~Mancini, S.~R. Bulo, E.~Ricci, and B.~Caputo, ``Modeling the
  background for incremental learning in semantic segmentation,'' in
  \emph{Proceedings of the IEEE/CVF Conference on Computer Vision and Pattern
  Recognition}, 2020, pp. 9233--9242.

\bibitem{badrinarayanan2017segnet}
V.~Badrinarayanan, A.~Kendall, and R.~Cipolla, ``Segnet: A deep convolutional
  encoder-decoder architecture for image segmentation,'' \emph{IEEE T-PAMI},
  vol.~39, no.~12, pp. 2481--2495, 2017.

\bibitem{ghiasi2016laplacian}
G.~Ghiasi and C.~C. Fowlkes, ``Laplacian pyramid reconstruction and refinement
  for semantic segmentation,'' in \emph{ECCV}, 2016.

\bibitem{ozdemir2018learn}
F.~Ozdemir, P.~Fuernstahl, and O.~Goksel, ``Learn the new, keep the old:
  Extending pretrained models with new anatomy and images,'' in
  \emph{International Conference on Medical Image Computing and
  Computer-Assisted Intervention}, 2018, pp. 361--369.

\bibitem{ozdemir2019extending}
F.~Ozdemir and O.~Goksel, ``Extending pretrained segmentation networks with
  additional anatomical structures,'' \emph{International journal of computer
  assisted radiology and surgery}, pp. 1--9, 2019.

\bibitem{tasar2019incremental}
O.~Tasar, Y.~Tarabalka, and P.~Alliez, ``Incremental learning for semantic
  segmentation of large-scale remote sensing data,'' \emph{IEEE Journal of
  Selected Topics in Applied Earth Observations and Remote Sensing}, vol.~12,
  no.~9, pp. 3524--3537, 2019.

\bibitem{michieli2019incremental}
U.~Michieli and P.~Zanuttigh, ``Incremental learning techniques for semantic
  segmentation,'' in \emph{ICCV-WS}, 2019, pp. 0--0.

\bibitem{de2019continual}
M.~De~Lange, R.~Aljundi, M.~Masana, S.~Parisot, X.~Jia, A.~Leonardis,
  G.~Slabaugh, and T.~Tuytelaars, ``Continual learning: A comparative study on
  how to defy forgetting in classification tasks,'' 2019.

\bibitem{shin2017continual}
H.~Shin, J.~K. Lee, J.~Kim, and J.~Kim, ``Continual learning with deep
  generative replay,'' in \emph{NeurIPS}, 2017.

\bibitem{hou2019learning}
S.~Hou, X.~Pan, C.~C. Loy, Z.~Wang, and D.~Lin, ``Learning a unified classifier
  incrementally via rebalancing,'' in \emph{CVPR}, 2019.

\bibitem{wu2018memory}
C.~Wu, L.~Herranz, X.~Liu, J.~van~de Weijer, B.~Raducanu \emph{et~al.},
  ``Memory replay gans: Learning to generate new categories without
  forgetting,'' in \emph{NeurIPS}, 2018.

\bibitem{ostapenko2019learning}
O.~Ostapenko, M.~Puscas, T.~Klein, P.~Jahnichen, and M.~Nabi, ``Learning to
  remember: A synaptic plasticity driven framework for continual learning,'' in
  \emph{CVPR}, 2019.

\bibitem{kirkpatrick2017overcoming}
J.~Kirkpatrick, R.~Pascanu, N.~Rabinowitz, J.~Veness, G.~Desjardins, A.~A.
  Rusu, K.~Milan, J.~Quan, T.~Ramalho, A.~Grabska-Barwinska \emph{et~al.},
  ``Overcoming catastrophic forgetting in neural networks,'' \emph{Proceedings
  of the national academy of sciences}, vol. 114, no.~13, pp. 3521--3526, 2017.

\bibitem{chaudhry2018riemannian}
A.~Chaudhry, P.~K. Dokania, T.~Ajanthan, and P.~H. Torr, ``Riemannian walk for
  incremental learning: Understanding forgetting and intransigence,'' in
  \emph{ECCV}, 2018.

\bibitem{zenke2017continual}
F.~Zenke, B.~Poole, and S.~Ganguli, ``Continual learning through synaptic
  intelligence,'' in \emph{ICML}, 2017.

\bibitem{dhar2019learning}
P.~Dhar, R.~V. Singh, K.-C. Peng, Z.~Wu, and R.~Chellappa, ``Learning without
  memorizing,'' in \emph{CVPR}, 2019.

\bibitem{mallya2018packnet}
A.~Mallya and S.~Lazebnik, ``Packnet: Adding multiple tasks to a single network
  by iterative pruning,'' in \emph{CVPR}, 2018.

\bibitem{mallya2018piggyback}
A.~Mallya, D.~Davis, and S.~Lazebnik, ``Piggyback: Adapting a single network to
  multiple tasks by learning to mask weights,'' in \emph{ECCV}, 2018.

\bibitem{rusu2016progressive}
A.~A. Rusu, N.~C. Rabinowitz, G.~Desjardins, H.~Soyer, J.~Kirkpatrick,
  K.~Kavukcuoglu, R.~Pascanu, and R.~Hadsell, ``Progressive neural networks,''
  2016.

\bibitem{wu2019large}
Y.~Wu, Y.~Chen, L.~Wang, Y.~Ye, Z.~Liu, Y.~Guo, and Y.~Fu, ``Large scale
  incremental learning,'' in \emph{CVPR}, 2019.

\bibitem{aljundi2018memory}
R.~Aljundi, F.~Babiloni, M.~Elhoseiny, M.~Rohrbach, and T.~Tuytelaars, ``Memory
  aware synapses: Learning what (not) to forget,'' in \emph{ECCV}, 2018.

\bibitem{fini2020online}
E.~Fini, S.~Lathuili{\`e}re, E.~Sangineto, M.~Nabi, and E.~Ricci, ``Online
  continual learning under extreme memory constraints,'' \emph{ECCV}, 2020.

\bibitem{shmelkov2017incremental}
K.~Shmelkov, C.~Schmid, and K.~Alahari, ``Incremental learning of object
  detectors without catastrophic forgetting,'' in \emph{ICCV}, 2017.

\bibitem{lee2019ficklenet}
J.~Lee, E.~Kim, S.~Lee, J.~Lee, and S.~Yoon, ``Ficklenet: Weakly and
  semi-supervised semantic image segmentation using stochastic inference,'' in
  \emph{Proceedings of the IEEE conference on computer vision and pattern
  recognition}, 2019, pp. 5267--5276.

\bibitem{huang2018weakly}
Z.~Huang, X.~Wang, J.~Wang, W.~Liu, and J.~Wang, ``Weakly-supervised semantic
  segmentation network with deep seeded region growing,'' in \emph{Proceedings
  of the IEEE Conference on Computer Vision and Pattern Recognition}, 2018, pp.
  7014--7023.

\bibitem{sun2020mining}
G.~Sun, W.~Wang, J.~Dai, and L.~Van~Gool, ``Mining cross-image semantics for
  weakly supervised semantic segmentation,'' in \emph{European conference on
  computer vision}.\hskip 1em plus 0.5em minus 0.4em\relax Springer, 2020.

\bibitem{dai2015boxsup}
J.~Dai, K.~He, and J.~Sun, ``Boxsup: Exploiting bounding boxes to supervise
  convolutional networks for semantic segmentation,'' in \emph{Proceedings of
  the IEEE international conference on computer vision}, 2015, pp. 1635--1643.

\bibitem{papandreou2015weakly}
G.~Papandreou, L.-C. Chen, K.~P. Murphy, and A.~L. Yuille, ``Weakly-and
  semi-supervised learning of a deep convolutional network for semantic image
  segmentation,'' in \emph{Proceedings of the IEEE international conference on
  computer vision}, 2015, pp. 1742--1750.

\bibitem{khoreva2017simple}
A.~Khoreva, R.~Benenson, J.~Hosang, M.~Hein, and B.~Schiele, ``Simple does it:
  Weakly supervised instance and semantic segmentation,'' in \emph{Proceedings
  of the IEEE conference on computer vision and pattern recognition}, 2017, pp.
  876--885.

\bibitem{tang2018normalized}
M.~Tang, A.~Djelouah, F.~Perazzi, Y.~Boykov, and C.~Schroers, ``Normalized cut
  loss for weakly-supervised cnn segmentation,'' in \emph{Proceedings of the
  IEEE Conference on Computer Vision and Pattern Recognition}, 2018, pp.
  1818--1827.

\bibitem{oh2017exploiting}
S.~J. Oh, R.~Benenson, A.~Khoreva, Z.~Akata, M.~Fritz, and B.~Schiele,
  ``Exploiting saliency for object segmentation from image level labels,'' in
  \emph{2017 IEEE conference on computer vision and pattern recognition
  (CVPR)}.\hskip 1em plus 0.5em minus 0.4em\relax IEEE, 2017, pp. 5038--5047.

\bibitem{ahn2018learning}
J.~Ahn and S.~Kwak, ``Learning pixel-level semantic affinity with image-level
  supervision for weakly supervised semantic segmentation,'' in
  \emph{Proceedings of the IEEE Conference on Computer Vision and Pattern
  Recognition}, 2018, pp. 4981--4990.

\bibitem{ahn2019weakly}
J.~Ahn, S.~Cho, and S.~Kwak, ``Weakly supervised learning of instance
  segmentation with inter-pixel relations,'' in \emph{Proceedings of the
  IEEE/CVF Conference on Computer Vision and Pattern Recognition}, 2019, pp.
  2209--2218.

\bibitem{araslanov2020single}
N.~Araslanov and S.~Roth, ``Single-stage semantic segmentation from image
  labels,'' in \emph{Proceedings of the IEEE/CVF Conference on Computer Vision
  and Pattern Recognition}, 2020, pp. 4253--4262.

\bibitem{chang2020weakly}
Y.-T. Chang, Q.~Wang, W.-C. Hung, R.~Piramuthu, Y.-H. Tsai, and M.-H. Yang,
  ``Weakly-supervised semantic segmentation via sub-category exploration,'' in
  \emph{Proceedings of the IEEE/CVF Conference on Computer Vision and Pattern
  Recognition}, 2020, pp. 8991--9000.

\bibitem{zhou2016cam}
B.~Zhou, A.~Khosla, A.~Lapedriza, A.~Oliva, and A.~Torralba, ``Learning deep
  features for discriminative localization,'' in \emph{Proceedings of the IEEE
  conference on computer vision and pattern recognition}, 2016, pp. 2921--2929.

\bibitem{selvaraju2017gradcam}
R.~R. Selvaraju, M.~Cogswell, A.~Das, R.~Vedantam, D.~Parikh, and D.~Batra,
  ``Grad-cam: Visual explanations from deep networks via gradient-based
  localization,'' in \emph{Proceedings of the IEEE international conference on
  computer vision}, 2017, pp. 618--626.

\bibitem{adams1994seeded}
R.~Adams and L.~Bischof, ``Seeded region growing,'' \emph{IEEE Transactions on
  pattern analysis and machine intelligence}, vol.~16, no.~6, pp. 641--647,
  1994.

\bibitem{srivastava2014dropout}
N.~Srivastava, G.~Hinton, A.~Krizhevsky, I.~Sutskever, and R.~Salakhutdinov,
  ``Dropout: a simple way to prevent neural networks from overfitting,''
  \emph{The journal of machine learning research}, vol.~15, no.~1, pp.
  1929--1958, 2014.

\bibitem{arbelaez2014multiscale}
P.~Arbel{\'a}ez, J.~Pont-Tuset, J.~T. Barron, F.~Marques, and J.~Malik,
  ``Multiscale combinatorial grouping,'' in \emph{Proceedings of the IEEE
  conference on computer vision and pattern recognition}, 2014, pp. 328--335.

\bibitem{rother2004grabcut}
C.~Rother, V.~Kolmogorov, and A.~Blake, ``" grabcut" interactive foreground
  extraction using iterated graph cuts,'' \emph{ACM transactions on graphics
  (TOG)}, vol.~23, no.~3, pp. 309--314, 2004.

\bibitem{tang2018regularized}
M.~Tang, F.~Perazzi, A.~Djelouah, I.~Ben~Ayed, C.~Schroers, and Y.~Boykov, ``On
  regularized losses for weakly-supervised cnn segmentation,'' in
  \emph{Proceedings of the European Conference on Computer Vision (ECCV)},
  2018, pp. 507--522.

\bibitem{wang2019boundary}
B.~Wang, G.~Qi, S.~Tang, T.~Zhang, Y.~Wei, L.~Li, and Y.~Zhang, ``Boundary
  perception guidance: a scribble-supervised semantic segmentation approach,''
  in \emph{IJCAI International Joint Conference on Artificial Intelligence},
  2019.

\bibitem{bharath2011sbd}
B.~Hariharan, P.~Arbelaez, L.~Bourdev, S.~Maji, and J.~Malik, ``Semantic
  contours from inverse detectors,'' in \emph{International Conference on
  Computer Vision (ICCV)}, 2011.

\bibitem{rota2018place}
S.~Rota~Bul{\`o}, L.~Porzi, and P.~Kontschieder, ``In-place activated batchnorm
  for memory-optimized training of dnns,'' in \emph{CVPR}, 2018.

\bibitem{he2016deep}
K.~He, X.~Zhang, S.~Ren, and J.~Sun, ``Deep residual learning for image
  recognition,'' in \emph{CVPR}, 2016.

\bibitem{xie2017aggregated}
S.~Xie, R.~Girshick, P.~Doll{\'a}r, Z.~Tu, and K.~He, ``Aggregated residual
  transformations for deep neural networks,'' in \emph{Proceedings of the IEEE
  conference on computer vision and pattern recognition}, 2017, pp. 1492--1500.

\bibitem{krahenbuhl2011efficient}
P.~Kr{\"a}henb{\"u}hl and V.~Koltun, ``Efficient inference in fully connected
  crfs with gaussian edge potentials,'' \emph{Advances in neural information
  processing systems}, vol.~24, pp. 109--117, 2011.

\bibitem{xian2019semantic}
Y.~Xian, S.~Choudhury, Y.~He, B.~Schiele, and Z.~Akata, ``Semantic projection
  network for zero-and few-label semantic segmentation,'' in \emph{Proceedings
  of the IEEE Conference on Computer Vision and Pattern Recognition}, 2019, pp.
  8256--8265.

\end{thebibliography}

\begin{IEEEbiography}
[{\includegraphics[width=1in,height=1.25in,clip,keepaspectratio]{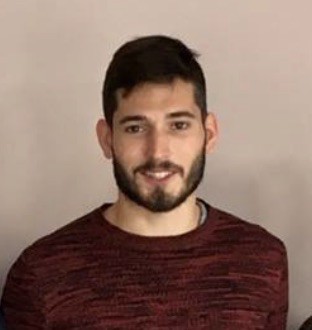}}]{Fabio Cermelli}
 is a Ph.D. student in Computer and Control Engineering at the Politecnico di Torino, funded by the Italian Insitute of Technology (IIT). He received his master thesis in Software Engineering (Computer Engineering) with honors at the Politecnico di Torino in 2018. He is member of the Visual Learning and Multimodal Applications Laboratory (VANDAL), supervised by Prof. Barbara Caputo. During his first year, he was a visiting Ph.D. student in the Technologies of Vision Laboratory at FBK.
\end{IEEEbiography}

\begin{IEEEbiography}
[{\includegraphics[width=1in,height=1.25in,clip,keepaspectratio]{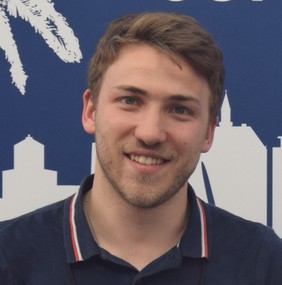}}]{Massimiliano~Mancini}
is a post-doctoral researcher at the Cluster of Excellence  in Machine Learning of the University of Tübingen, in the Explainable Machine Learning group, lead by Prof. Zeynep Akata. He completed his PhD in Engineering in Computer Science at the Sapienza University of Rome in 2020. 
During the Ph.D. he has been a member of the ELLIS PhD program, 
the Technologies of Vision lab at Fondazione Bruno Kessler, and the Visual Learning and Multimodal Applications Laboratory of the Italian Institute of Technology. His research interests include transfer learning across domains and learning from low supervision.
\end{IEEEbiography}

\begin{IEEEbiography}
[{\includegraphics[width=1in,height=1.25in,clip,keepaspectratio]{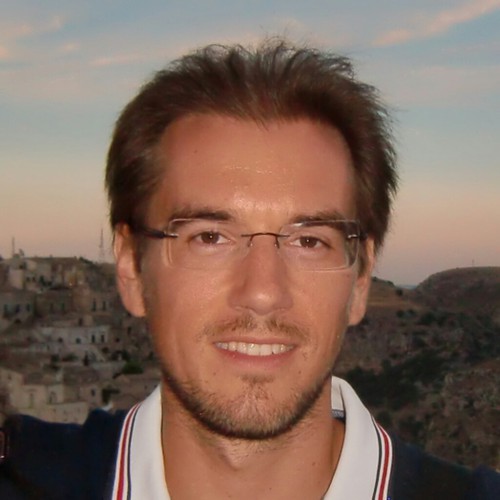}}]{Samuel~Rota~Bul\'o}
received the PhD in computer science at University of Venice in 2009. He worked there as PostDoc and held teaching positions until 2013. Then, he became a researcher for FBK in computer vision and machine learning. In 2017, he moved to Mapillary Research, where he worked as a senior researcher. He was awarded the prestigious Marr Prize in 2015. He serves on the editorial board for "Pattern Recognition" and “International Journal of Machine Learning and Cybernetics”  and is regularly on the program committee of international conferences of his field. He participated to several  EU projects (SIMBAD, VENTURI, REPLICATE).
\end{IEEEbiography}

\begin{IEEEbiography}
[{\includegraphics[width=1in,height=1.25in,clip,keepaspectratio]{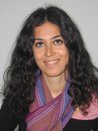}}]{Elisa~Ricci}
 is an associate professor at University of Trento and a researcher at Fondazione Bruno Kessler. She received her PhD from the University of Perugia in 2008. She has since been a post-doctoral researcher at Idiap Research Institute and an assistant professor at University of Perugia. She was also a visiting researcher at University of Bristol. Her research interests
are mainly in the areas of computer vision and machine learning.
\end{IEEEbiography}

\begin{IEEEbiography}
[{\includegraphics[width=1in,height=1.25in,clip,keepaspectratio]{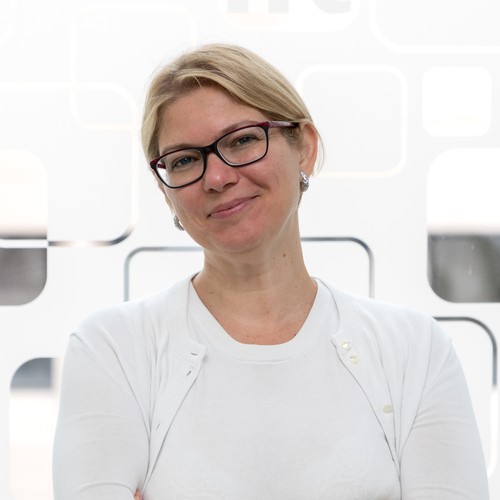}}]{Barbara~Caputo}
 is Full Professor at the  DAUIN Department of Control and Computer Engineering of Politecnico di Torino and Principal Investigator at the Italian Institute of Technology (IIT), where she leads the Visual Learning and Multimodal Applications Laboratory (VANDAL).  Her main research interest is to develop algorithms for learning, recognition and categorization of visual and multimodal patterns for artificial autonomous systems. These features are crucial to enable robots to represent and understand their surroundings, to learn and reason about it, and ultimately to equip them with cognitive capabilities. Her research is sponsored by the Swiss National Science Foundation (SNSF), the Italian Ministry for Education, University and Research (MIUR), the European Commission (EC) and the European Research Council (ERC).
\end{IEEEbiography}

\end{document}